\pdfoutput=1
\documentclass[11pt]{article}

\usepackage{bbold}
\usepackage{bbm}
\usepackage[final]{acl}
\usepackage{listings}
\usepackage{footmisc}
\usepackage{sidecap}
\usepackage{enumitem}
\usepackage{algorithm}
\usepackage{algpseudocode}

\usepackage{xcolor}  
\usepackage{booktabs}       
\usepackage{array}          
\usepackage{longtable}      
\usepackage{graphicx}       
\usepackage{subcaption}     
\usepackage{caption}
\usepackage{wrapfig}        
\usepackage{geometry}       
\usepackage{tabularx}
\usepackage{colortbl}

\definecolor{lightgray}{gray}{0.9}

\usepackage[most]{tcolorbox}

\usepackage{amsmath}        

\usepackage{times}
\usepackage{latexsym}
\usepackage{multirow}       

\usepackage[utf8]{inputenc}
\usepackage[OT1]{fontenc}

\usepackage{inconsolata}
\usepackage{adjustbox}
\usepackage{microtype}
\usepackage{pifont}
\usepackage{fontawesome5}
\usepackage[misc]{ifsym}
\usepackage{ragged2e}

\newcommand{\cmark}{\textcolor{green}{\ding{51}}}  
\newcommand{\xmark}{\textcolor{red}{\ding{55}}}    
\newcommand{\name}{EnterpriseBench}

\usepackage{hyperref}

\title{Can LLMs Help You at Work?  A Sandbox for \newline Evaluating LLM Agents in Enterprise Environments}
\usepackage{ragged2e} 

\title{
  \begin{tabular}{@{}c@{\hspace{1em}}p{0.75\textwidth}@{}}
    \raisebox{-.5\height}{\includegraphics[height=3em]{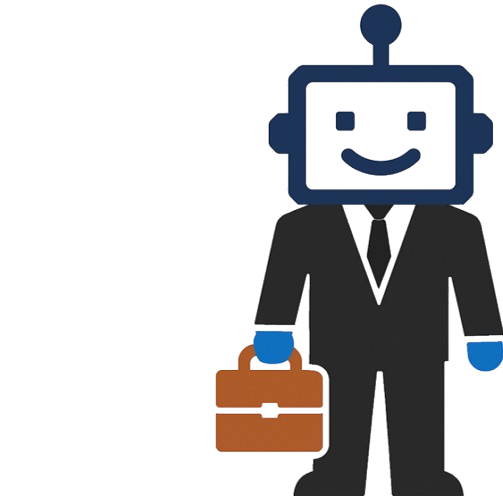}} &
    \justifying
    Can LLMs Help You at Work? 
    A Sandbox for \newline Evaluating LLM Agents in Enterprise Environments
  \end{tabular}
}

\usepackage{hyperref}  

\author{
  \textbf{Harsh Vishwakarma}\thanks{Equal contribution as co-first authors.} \quad
  \textbf{Ankush Agarwal}\footnotemark[1] \quad
  \textbf{Ojas Patil} \\
  \textbf{Chaitanya Devaguptapu} \quad
  \textbf{Mahesh Chandran} \\
  Fujitsu Research \\
    \href{https://ast-fri.github.io/EnterpriseBench}{\textit{EnterpriseBench - Tech Blog}} \\
  \texttt{\{harsh.vishwakarma, ankush.agarwal\}@fujitsu.com}
}

\begin{document}
\maketitle

\renewcommand{\thefootnote}{}
\footnotetext{\href{https://github.com/ast-fri/EnterpriseBench.git}{Code}    \href{https://huggingface.co/datasets/AST-FRI/EnterpriseBench}{Data}}
\renewcommand{\thefootnote}{\arabic{footnote}}

\begin{abstract}
Enterprise systems are crucial for enhancing productivity and decision-making among employees and customers. Integrating LLM based systems into enterprise systems enables intelligent automation, personalized experiences, and efficient information retrieval, driving operational efficiency and strategic growth. However, developing and evaluating such systems is challenging due to the inherent complexity of enterprise environments, where data is fragmented across multiple sources and governed by sophisticated access controls. We present \name, a comprehensive benchmark that simulates enterprise settings, featuring 500 diverse tasks across software engineering, HR, finance, and administrative domains. Our benchmark uniquely captures key enterprise characteristics including data source fragmentation, access control hierarchies, and cross-functional workflows. Additionally, we provide a novel data generation pipeline that creates internally consistent enterprise tasks from organizational metadata. Experiments with state-of-the-art LLM agents demonstrate that even the most capable models achieve only 41.8\% task completion, highlighting significant opportunities for improvement in enterprise-focused AI systems. 

\end{abstract}
\vspace{-0.2cm}
\section{Background and Introduction}\label{sec:intro}
\begin{figure*}[htbp]
    \centering
    \includegraphics[width=0.9\textwidth]{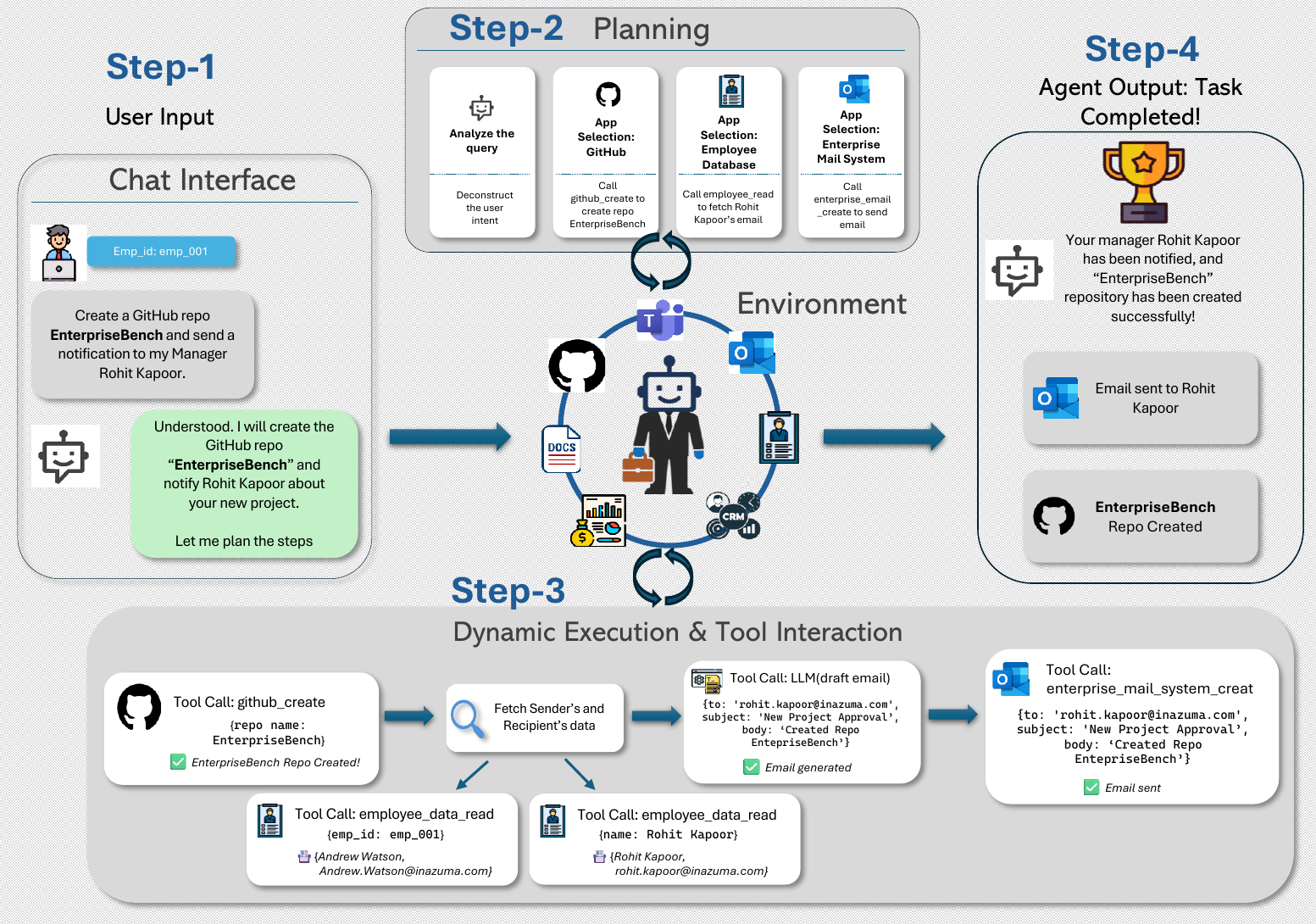}
    \caption[Task Execution in EnterpriseBench.]{\textbf{Task Execution in EnterpriseBench.} This figure illustrates how an LLM-based agent interacts with the enterprise environment. Given a task, the agent perceives the available enterprise tools, applications, and data sources, formulates a reasoning plan, and executes actions to complete the task.}
    \label{fig:E-Bench}
    \vspace{-0.4cm}
\end{figure*}

Large Language Models (LLMs) are fundamentally transforming how enterprises operate, driving improvements in productivity across departments \cite{Google_2024_LLM_News, Meta_2024_LLM_News, Nicholas_2024}.  These models have demonstrated remarkable capabilities in automating knowledge-intensive tasks, from question answering and code generation to report writing and data analysis~\cite{brachman2024knowledge, jiang2024enhancing, GitHubCopilot}. Recent advancements have led to emergence of Compound AI Systems (CAI)~\cite{compound-ai-blog, lin2024llm} (also  referred to as Agents~\cite{LangChain_2024, Anthropic_2024}) that can orchestrate complex workflows for solving various tasks. These systems, exemplified by tools like Devin~\cite{CognitionLabs2024Devin} and Glean~\cite{Glean}, can automatically search across information sources, analyze data, and even initiate actions when human intervention is needed.

\noindent However, developing effective CAI systems for enterprises faces a critical challenge: enterprise data is inherently complex and fragmented across multiple sources, including email systems, Customer Relationship Management (CRM) platforms, SharePoint sites, internal wikis, and ticketing systems. This fragmentation is further complicated by sophisticated access control mechanisms that govern who can access specific information resources. Even seemingly simple queries often require orchestrating data gathering from multiple sources, executing database calls, and performing complex reasoning across diverse information types. While current research has made progress in developing CAI systems for specific use-cases relevant to enterprises, the unique challenges of enterprise environments—particularly around data fragmentation and access control—remain largely unaddressed with current CAI systems.

\vspace{-0.1cm}
\noindent  To illustrate challenges and complexities of the CAI, consider an enterprise specific scenario:  an employee asks, \textit{"Create a GitHub repository named EnterpriseBench and generate a notification message to my manager informing him about the repository creation."} This seemingly straightforward request requires a complex workflow that traditional approaches like Retrieval-Augmented Generation (RAG)~\cite{bruckhaus2024rag} and existing LLM agents~\cite{talebirad2023multi, zhang2024exploring, li2019multi} struggle to handle. A robust enterprise-specific CAI system must orchestrate multiple subtasks for this: create the GitHub repository EnterpriseBench, resolve the sender and recipient details, generate a formal notification message—all while respecting access controls and organizational hierarchies. These requirements highlight the need for sophisticated CAI systems that can (1) integrate multiple enterprise data sources and tools, (2) enforce access controls, (3) coordinate multiple tasks, and (4) maintain context across system interactions (as shown in Figure ~\ref{fig:E-Bench}).

\vspace{-0.1cm}
\noindent To enable development of such systems, we introduce \name, the first comprehensive benchmark that simulates the data from enterprise environments. By providing a benchmark that mirrors 
complexities of real-world scenarios without using sensitive real data, \name \space enables rapid prototyping and evaluation of CAI systems for enterprise settings. This allows organizations to validate and refine their CAI systems before deploying them on actual enterprise data. Our dataset spans multiple domains, including Software Engineering (code repositories, documentation), Sales and CRM (customer interactions), Finance (budgets, expense reports), IT support (ticketing systems, incident reports), HR (policies, employee records), and Internal Communication platforms (simulated team and email conversations). \name \space emphasizes persona-based tasks that require adherence to access controls and organizational hierarchies. Additionally, we also introduce an automated task creation framework that generates complex, multi-source tasks conditioned on persona roles and enterprise constraints.

\noindent We conduct a comprehensive evaluation of five large language models, including GPT-4o \cite{hurst2024gpt}, Claude 3.5 \cite{anthropic2024claude}, O1-mini \cite{openai_o1_mini}, LLaMA \cite{touvron2023llama}—to assess their ability to generate complete plans for accomplishing a given task. Our evaluation spans four planning strategies, including ReAct \cite{yao2022react} and Chain-of-Thought (CoT) \cite{wei2022chain}, implemented using two different frameworks, LangChain \cite{LangChain_2024} and DSPy \cite{khattab2024DSPy}.
Our key contributions are listed below.
\noindent
\begin{itemize}[leftmargin=*,nosep]
\setlength\itemsep{0em}
    \item A comprehensive benchmark of 500 enterprise tasks across IT, HR, Sales and Finance, featuring multi-step reasoning, access controls, and cross-functional workflows.
    \item Our comprehensive evaluations shows a significant performance gap in current CAI systems, with even state-of-the-art models achieving only 41.8\% task completion.
    \item A simulated enterprise sandbox environment is created for benchmark development, comprising data domains such as chat systems, emails, and code workflows, along with representative employee information aligned with these domains.
    \item A persona-based task framework that generates contextually appropriate challenges, testing both technical capabilities and organizational constraints.
\end{itemize}

\vspace{-0.2cm}
\section{Related Work}

\vspace{-0.2cm}
\textbf{Compound AI Systems}
LLMs have emerged as powerful tools, demonstrating excellence in tasks such as processing and generating human-like text \cite{team2023gemini, achiam2023gpt}, writing code \cite{chen2021evaluating}, and performing complex reasoning \cite{khetan2020causal}. Beyond these fundamental capabilities, LLMs show immense potential within Compound AI Systems, enabling collaborative problem-solving, dynamic interactions, and advanced decision-making \cite{yao2022react, xi2023rise, wei2022chain}. As tasks grow in complexity and scope, leveraging multiple LLMs in a cooperative framework becomes a natural strategy to enhance their effectiveness. To evaluate these systems, specialized benchmarks are developed, which are discussed in the next module.


\noindent \textbf{Evaluation of Compound AI System}
Compound AI Systems have been developed to address a wide range of tasks, including scientific experimentation \cite{ghafarollahi2024protagents, boiko2023emergent, m2024augmenting}, embodied intelligence \cite{brohan2023can}, societal simulations \cite{gao2023s, li2023large}, and web-based environemnts such as Mind2Web \cite{dengmind2web}, WebArena \cite{zhouwebarena}, and WebShop \cite{yao2022webshop}. 
Recently, benchmarks have begun to emerge for more specialized settings, such as software engineering \cite{jimenezswe, li2024devbench}, computing environments \cite{xie2024osworld, bonatti2024windows}, workplace \cite{stylesworkbench}, text-to-SQL workflows \cite{leispider}, and real-world task planning \cite{yaotau, liuagentbench, xie2024travelplanner}.
Despite these advancements, there remains a significant gap in the development of enterprise-simulated environments that reflect real-world, day-to-day business operations. The closest efforts in this direction such as \citet{xu2024theagentcompany, huang-etal-2025-crmarena} focus on narrow domains like database management or CRM systems. However, none of them address the challenges of managing large volumes of data spread across diverse domains, formats, and systems—a key requirement for evaluating Compound AI Systems in realistic enterprise settings.

To address this gap, we propose a novel benchmark, \name, specifically designed for enterprise scenarios. This benchmark offers a robust framework for evaluating LLM-based agents under realistic, domain-relevant conditions, thereby supporting the development of effective and reliable enterprise AI systems. A comparison with other related benchmarks is presented in Table \ref{tab:benchmark_comparison}.

\vspace{-0.2cm}
\section{\name: Crafting a Simulated Enterprise Benchmark}



We have developed an enterprise sandbox environment that simulates a realistic company setting. This environment includes synthetic company data enriched with employee-specific details such as chat logs, emails, and GitHub activity. The data sources are constructed by gathering publicly available information from the internet and applying rule-based processing techniques, guided by domain experts to ensure authenticity. Based on this simulated data, a variety of enterprise tasks are generated within the sandbox, with strict access control policies in place to support secure and realistic interactions.

The subsequent sections elaborate on the key components of our benchmark. Section~\ref{sec: ent_tasks} outlines the design of enterprise tasks. Section~\ref{sec: sandbox_env} details the simulation of the enterprise sandbox, followed by the automatic task construction pipeline described in Section~\ref{sec: task_construction}. Section~\ref{sec: api_calls} presents the API calls and functions implemented within the sandbox to support LLM agents. Finally, Section~\ref{sec: exp_studies} reports an expert study conducted to assess the realism and validity of the sandbox environment and tasks.

\vspace{-0.2cm}
\subsection{\name Tasks}
\label{sec: ent_tasks}
\newcolumntype{P}[1]{>{\centering\arraybackslash}p{#1}} 

\newcolumntype{Y}{>{\centering\arraybackslash}X} 

\begin{table*}[t]
\centering
\small
\begin{tabularx}{\textwidth}{
    >{\centering\arraybackslash}m{2.0cm}   
      >{\centering\arraybackslash}m{4.5cm}                                  
    >{\centering\arraybackslash}m{2.0cm}  
    >{\centering\arraybackslash}m{2.0cm}  
    >{\centering\arraybackslash}m{3.0cm}                                    }
\toprule
\textbf{Domain} & \textbf{Task Description} & \textbf{User Employee} & \textbf{Task Category} & \textbf{Tools} \\
\midrule
SWE & Create a GitHub repo EnterpriseBench and notify my Manager Rohit Kapoor. 
& emp\_001 (Level-10) & CRUD & github\_create, enterprise\_mail\_system\_create, employee\_data\_read \\

IT Management & Can you show me the ticket IDs for urgent issues I'm involved with? 
& emp\_1106 (Level-10) & Search & it\_service\_management\_read \\

Business Development & Can you help me find the thread for my recent conversations with Julian? I need it for follow-up. 
& emp\_1180 (Level-12) & Search & conversations\_read \\

Employee Database Management & Can you provide a breakdown of my total leaves taken so far this year? 
& emp\_0726 (Level-09) & Search & employee\_data\_read \\

HR & Send a mail to Rahul Khanna to notify the employees regarding new PoSH policy document. 
& emp\_0653 (Level-09) & CRUD & employee\_data\_read, enterprise\_mail\_system\_create \\
\bottomrule
\end{tabularx}
\caption{Examples of \name \space tasks across domains, categorized by task type and tools.}
\label{tab:enterprise_task_examples}
\end{table*}

\begin{figure}
    \centering
    \includegraphics[width=\columnwidth]{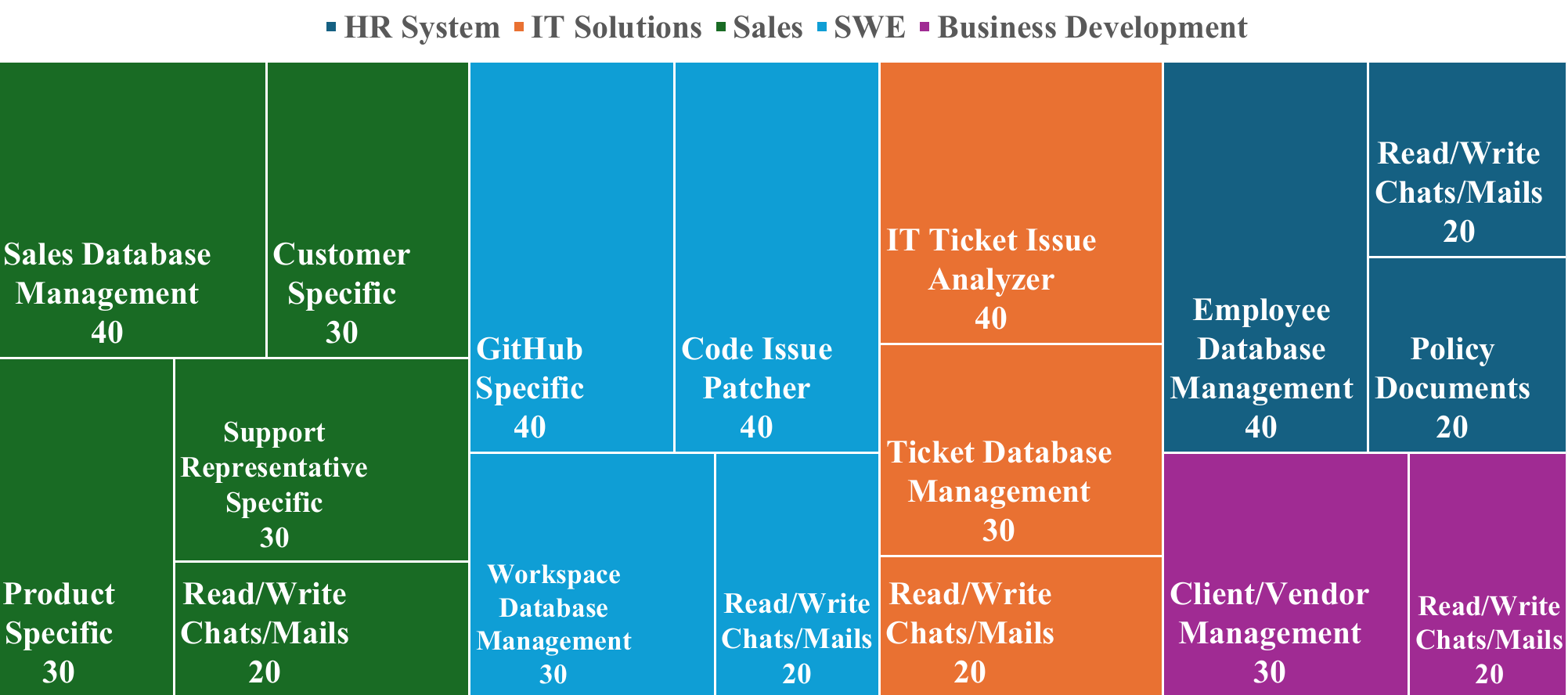}
    \caption{Classification of Tasks by Domain (counts)}
    \label{fig:Task_classification}
\end{figure}



Our benchmark includes 500 enterprise tasks spanning five major domains: Human Resources (HR), Information Technology (IT), Software Engineering (SWE), Business Operations, and Sales. Each task is carefully designed to assess the capabilities of Compound AI systems in enterprise setting. 
To capture a broad range of functionalities, the tasks are grouped into three primary categories: search tasks, CRUD (Create, Read, Update, Delete) tasks, and unanswerable, which account for 65\%, 30\%, and 5\% of the benchmark, respectively.
The domain-wise distribution of tasks is shown in Figure~\ref{fig:Task_classification}, and the average task complexity is defined by the number of tools required to solve a task, which is 3. Representative examples of tasks included in the benchmark are shown in Table~\ref{tab:enterprise_task_examples}.

\subsection{\name \space Sandbox: Simulating Enterprise Data and Roles}
\label{sec: sandbox_env}

\begin{table}[h!]
\centering
\small
\arrayrulecolor{black} 
 \begin{tabular}{p{2.5cm} >{\centering\arraybackslash}m{3cm} >{\centering\arraybackslash}m{1cm}}
 \toprule
\textbf{Application / Data Source} & \textbf{Content Description} & \textbf{\# Instances}  \\
\midrule
\arrayrulecolor{lightgray}
\raisebox{-0.2\height}{\includegraphics[width=0.35cm]{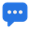}}~Chats & Conversations between employees & 3000\\ \hline
\raisebox{-0.2\height}{\includegraphics[width=0.35cm]{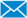}}~Enterprise Mail System & Internal and external email threads & 4500\\ \hline
\raisebox{-0.2\height}{\includegraphics[width=0.35cm]{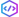}}~Code Workspace & Source code, issues, issue git patch & 1000\\ \hline
\raisebox{-0.2\height}{\includegraphics[width=0.35cm]{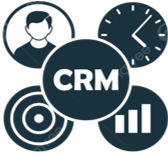}}~Customer Relationship Management & Records on sales, product, customers, support chats, sentiment data, invoices, and purchase orders & 30195\\ \hline
\raisebox{-0.2\height}{\includegraphics[width=0.35cm]{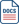}}~Enterprise Policy Documents & Policy and compliance documents & 24\\ \hline
\raisebox{-0.2\height}{\includegraphics[width=0.35cm]{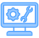}}~IT Service Management & Support tickets and issue logs & 163\\ \hline
\raisebox{-0.2\height}{\includegraphics[width=0.35cm]{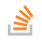}}~Enterprise Internal Overflow & Similar to Stack Overflow & 5000\\ \hline
\raisebox{-0.2\height}{\includegraphics[width=0.35cm]{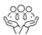}}~Enterprise Social Blog & Internal blog posts, company news and updates & 1000\\ \hline
\raisebox{-0.2\height}{\includegraphics[width=0.35cm]{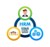}}~HR Management System & Employee records, resumes and organizational structure & 1260\\ \hline
\raisebox{-0.2\height}{\includegraphics[width=0.35cm]{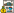}}~Business and Management & Client Details, Vendor Details, & 800 \\
\arrayrulecolor{black}\bottomrule 
\end{tabular}
\caption{Details of Data Sources/Applications in the \name \space Simulated Sandbox Environment}
\label{tab:simulation_environment}
\end{table}

\label{sec: method}

\noindent The enterprise sandbox environment is developed with careful consideration of three key components: \textit{Departments to Populate}, \textit{Data Sources to Collect}, and \textit{Compiling the Data to create the Simulation Environment}. We integrate both collected and synthetically generated data across multiple domains-HR, IT, Sales, Finance, and Software Development within a simulated organizational setting. Table \ref{tab:simulation_environment} show details regarding the data sources in \name.

\noindent Employee data is sourced from \citet{ayoobi2023looming}, filtered to include only relevant departments. To reflect organizational structure, employees are categorized into four roles-Associates, Team Leads, Managers, and Directors distributed in a 4:3:2:1 ratio per department. Additional attributes such as salary, leave records, and joining dates are introduced to mimic real-world enterprise dynamics.

\subsubsection{Sandbox Data Simulation}

The data simulation strategy is based on two primary methodologies.

\noindent \textbf{Leveraging the Collected Data} We collect the enterprise-related data from different sources (details in Table \ref{tab:enterprise-stats}) and use it to simulate the sandbox environment. Below, we explain how it is utilized.

\begin{itemize}[left=0pt, labelsep=0.5em, itemsep=0pt, topsep=0pt]
    \item \textit{Data Source Coverage:} Domain experts (details in Appendix \ref{sec:exp_study}) identified essential data sources for each department. For example, the Sales department should include Customer Support Chats, Product Sentiment Data, Product, Customer, Sales Data, Invoices, Purchase Orders, and Shipping Records.

    \item \textit{Pre-processing:}  
Collected data undergoes structural preprocessing, including extraction of entities (e.g., products, customers) and generation of contextual data (e.g., support conversations).

 \item \textit{Mapping to Employee Personas:}  
Data entries are linked to employee personas based on experience, skills, and roles. For instance, customer resolutions are semantically mapped to specific support personnel.

\item \textit{Enterprise Rephrasing:}  
Entries are rephrased using enterprise-specific metadata to ensure contextual consistency and realism.

\end{itemize}


\noindent \textbf{Generating Conversations and Emails}

\noindent Following the methodology in \citet{xu2024hr}, realistic conversations and emails are generated and grounded in curated data to reflect authentic enterprise communication. More details on generation are available in Appendix \ref{sec:sandbox}.

\vspace{0.1cm}
\subsubsection{Access Control Simulation}

To emulate enterprise-level security, we implement a dynamic Role-Based Access Control system, where permissions are assigned based on organizational role levels (specifically, Levels 9 through 14), task requirements, data sensitivity, and cross-departmental relationships. For example, enterprise social platforms are accessible to all employees, while access to internal repositories (such as GitHub) is restricted to designated technical teams and their management chain. Access control policies are initially generated with assistance from a LLM and subsequently validated by human experts. 

\subsection{\name \space Task Generation Pipeline}
\label{sec: task_construction}


We designed an LLM-based task generation pipeline to produce structured, high-quality tasks that require access to relevant data sources and tools, while also enforcing persona-specific access controls. The pipeline comprises four key stages: a) selecting the initial domain and persona for the task, b) selection from expert curated goal templates, c) generating the corresponding task based on the selected context, and d) refining the task iteratively. A stepwise explanation is provided below.

\subsubsection{Domain and Persona Selection}
We begin by
\vspace{-0.2cm}
\noindent\begin{itemize}[noitemsep, leftmargin=*]
    \item \textit{Task Domain Selection:} Among the available domains such as HR, IT, we randomly select a target domain for which task has to be generated.

    \item \textit{Persona Sampling:} From a set of personas curated by domain experts for each domain, a representative persona is sampled for the selected domain to serve as a proxy for task contextualization.
    
    \item \textit{Context Retrieval:} From the prepared data sources available in the sandbox environment, relevant contextual information associated with the sampled persona and domain is retrieved to ground the task in a realistic enterprise scenario.
    
\end{itemize}

\subsubsection{Expert Curated Goal Templates}

Creating generalizable goal templates across departments is inherently challenging due to the diversity and specificity of enterprise tasks. To address this, we leverage the O*NET 29.2 \footnote{\href{https://www.onetcenter.org/database.html}{O*NET 29.2}} release~\cite{rounds1999development}, a comprehensive taxonomy of occupations and task definitions developed by the U.S. Department. We manually curated goal templates (examples in Table \ref{tab:goal-templates}) tailored to departmental tasks, refining them through iterative reviews by domain experts to ensure contextual relevance and practical applicability. 

\subsubsection{Task Generation}

We use the persona and domain-relevant context, along with the selected goal template and available tools, to initiate the task generation process using LLM calls. We begin by
\vspace{-0.2cm}
\noindent\begin{itemize}[noitemsep, leftmargin=*]
    \item \textit{Entity Extraction:} Filters are applied on the persona-specific context to structure the input, reducing token count and enhancing the precision of downstream processing. This structured representation improves task grounding by highlighting salient information.
    \item \textit{Subgoal Decomposition:} The expert curated high-level goal is decomposed into fine-grained subgoals, including retrieval steps and action plans, by prompting the language model to operate in a closed, tool-aware environment. This stage introduces modularity into the task planning process.
    \item \textit{Task Structure:} Based on the subgoals and extracted entities, task structure is defined that can be mapped with the context entities. The structure mirror the reasoning sequence or plan that a compound AI system would follow to execute the complete task.
    \item \textit{Final Task Generation:} The final task is assembled by synthesizing the goal, subgoals, entities, and task structure, resulting in a fully formed, executable task representation.
\end{itemize}

\noindent A comprehensive description of how the ground truth is established can be found in Appendix~\ref{sec:add_res}.

\subsubsection{Iterative Improvement}
 Inspired by the iterative refinement method proposed in \citet{kim2023tree,yaotau}, a validation and rephrasing loop is applied. The generated task and ground truth is iteratively revised until it passes a checklist of validation criteria designed by human experts, ensuring clarity, feasibility, and alignment with task objectives. 
\\\\
\noindent We provide the end-to-end task generation procedure in Algorithm~\ref{alg:generate_emp_task}, included in Appendix~\ref{sec:add_res}. The LLM prompts used for task generation are detailed in Appendix~\ref{sec:prompts}. Information about the domain experts involved in designing goal templates, filtering ambiguous tasks, and other aspects of task generation is provided in Appendix~\ref{sec:exp_study}.

\subsection{Tools: API and Functions}
\label{sec: api_calls}

\name \space incorporates a suite of tools and functions designed to simulate enterprise operations across diverse domains (see Table~\ref{tab:app_tools} for the tool inventory). For search-based tasks, agents use domain-specific application interface tools that return results based on employee ID or semantic matching. CRUD-based tasks leverage create, read, update, and delete operations for each data source, enabling dynamic data manipulation. To reflect enterprise settings, tool and function outputs are regulated by an access control mechanism that enforces permission constraints.

\subsection{Expert Study}
\label{sec: exp_studies}
To ensure the correctness, realism, and practicality of \name, we conducted a user study involving ten experts from diverse professional backgrounds. The experts were selected through a Microsoft Form circulated internally; additional details on form and experts are provided in Appendix \ref{sec:exp_study}. The selection was based on their domain expertise to ensure relevant and informed feedback.
During the study, participants were first introduced to the sandbox environment, followed by a set of questions designed to assess their understanding of the setup. They were then presented with search-based and CRUD-based tasks and asked to find answers or perform the required operations. This step helped assess the correctness of the tasks.
Subsequently, participants evaluated the realism of each task using a scale ranging from “very unrealistic” to “very realistic,” and provided justifications for their ratings. Based on the scores, we applied a filtering process to retain only those tasks that met enterprise-specific quality standards. As a result, 80\% of the tasks were considered suitable, meaning they were correct, realistic, and aligned with enterprise applications, while the remaining tasks were discarded.

\vspace{-0.2cm}
\section{Experimental Setup}
\label{sec: exp_and_analysis}

\subsection{Enterprise LLM Agent Setup}
    To efficiently solve our enterprise search tasks, we design an LLM-based agent that follows a structured multi-step approach. Given a primary goal or task \( T \), the agent creates a plan by decomposing it into sub-goals or sub-tasks \( P = \{p_1, p_2, \dots, p_n\} \) using a reasoning-based method. These sub-goals are then refined into well-defined, solvable steps \( S = \{s_1, s_2, \dots, s_n\} \). The agent, defined as \( \mathcal{A} = f(\Theta, \mathcal{K}) \), where \( \Theta \) and \( \mathcal{K} \) are model parameters and prior knowledge, selects the appropriate tools or API functions \( T \) to optimize information retrieval and processing. It then iteratively executes each sub-task, constructing the final answer \( A \) or executing the final task. 

\noindent This setup ensures reliable execution of \name \space tasks by leveraging LLM Agents for multi-step reasoning, tool utilization, and execution.

\vspace{-0.2cm}
\subsection{Experimental Settings}
This section outlines our experimental setup, detailing the experimental data, baseline methods used to evaluate our benchmark, the evaluation metrics employed, and the implementation specifics.

\subsubsection{Experimental Dataset}
We conduct experiments by building CAI systems with a range of models. For evaluation with LangChain and DSPy, we use the benchmark of 500 samples. For supervised fine-tuning (SFT), we expand the dataset to 1k samples using our task generation pipeline and split it into training and test sets with a 4:1 ratio. For Direct Preference Optimization (DPO) \cite{rafailov2023direct}, we conduct experiment by creating 1200 preference pairs from SFT training examples, following the procedure described in OS-Genesis \cite{sun2024genesis}. The dataset formats for SFT and DPO are illustrated in Listings~\ref{lst:sft} and~\ref{lst:dpo}, respectively in Appendix.

\subsubsection{Baseline Methods}
\label{sec:baseline}

To evaluate the performance on the \name \space benchmark, we conducted experiments using several state-of-the-art models: \texttt{GPT-4o}\footnote{\label{gpt4o}\href{https://platform.openai.com/docs/models\#gpt-4o}{https://platform.openai.com/docs/models\#gpt-4o}}, \texttt{o1-mini}\footnote{\label{o1mini}\href{https://platform.openai.com/docs/models\#o1}{https://platform.openai.com/docs/models\#o1}} (via Azure AI Foundry), \texttt{Anthropic Claude 3.5-Sonnet}\footnote{\label{claude}\href{https://aws.amazon.com/bedrock/claude/}{https://aws.amazon.com/bedrock/claude/}}  (\texttt{anthropic.claude-3-5-sonnet-20240620-v1:0}) from Amazon Bedrock, as well as \texttt{Llama-3.1-8B}, and \texttt{Llama-3.3-70B}, also accessed via Amazon Bedrock.
Building on these models, CAI system baselines using a variety of planning strategies: no planning, Chain-of-Thought (CoT) reasoning \cite{wei2022chain}, ReAct-style reasoning \cite{yao2022react}, and goal-aware planning. To implement these systems, we adapted state-of-the-art agent frameworks, namely LangChain \cite{LangChain_2024} and DSPy \cite{khattab2024DSPy}. For DSPy, we employ an optimization-based few-shot prompting approach, while for LangChain, we provide two few-shot examples with each LLM call. Each system is designed to decompose primary goals into subgoals, select relevant data sources and tools, verify access controls, and execute tasks in an end-to-end manner, ensuring alignment with enterprise-specific requirements.

\vspace{-0.2cm}
\subsubsection{Implementation Details}

\noindent Experiments were conducted using two NVIDIA GPUs (80 GB each) for SFT and DPO training. Additional 8 GB GPUs were employed to load retrievers such as Colpali for implementing the EnterpriseBench environment, while LLM inference was carried out through APIs.
\noindent \begin{itemize}[leftmargin=*,nosep]
    \item \textit{Data Simulation}: We utilized \texttt{GPT-4o}\footref{gpt4o} to generate and rephrase all components of \name, ensuring consistency and high-quality data synthesis.
    \item \textit{Task Generation}: The task generation process was conducted using \texttt{GPT-4o}\footref{gpt4o}, implementing an end-to-end pipeline. Additionally, \texttt{Anthropic Claude 3.5-Sonnet}\footref{claude} was employed for final quality assessment of the generated tasks. It took approximately 1 minutes and 20 seconds to generate a single task.
    \item \textit{Tool Dependency and Execution}: Tool dependencies were defined using a structured \texttt{JSON} file containing detailed descriptions of all tools within \name. For tool execution, API calls were made to invoke various external tools. Further details on tool specifications and implementations can be found in Table \ref{tab:app_tools}.
    \item \textit{Context Retrieval}: We implemented \textit{id} based context retriever for text-based structured data, Colpali \cite{faysse2024colpali} for PDF documents, and query-to-SQL retrievers inspired by~\cite{zhang2025murre} for tabular content.
    
    \item \textit{SFT+DPO}: We implemented SFT using LoRA \cite{hu2022lora}, targeting the modules \texttt{q\_proj}, \texttt{k\_proj}, \texttt{v\_proj}, and \texttt{o\_proj}. All other hyperparameters followed the default \texttt{LoraConfig} in the TRL library from Hugging Face\footnote{\url{https://huggingface.co/docs/trl/index}}. DPO was implemented using the \texttt{DPOTrainer} from TRL with the same hyperparameters as SFT.

\noindent The hyperparameter configurations for LLM API calls and retrievers are summarized in Table~\ref{tab:api_hyperparams} and Table~\ref{tab:retriever_hyperparams} in Appendix.


\end{itemize}

\vspace{-0.2cm}
\subsection{Evaluation Metric}
\label{sec:evaluation_metric}

To evaluate Compound AI systems on \name, we assess the correctness of the final execution of each task. For all tasks, correctness is determined using Prometheus-2\footnote{\href{https://docs.llamaindex.ai/en/latest/examples/cookbooks/prometheus2_cookbook/}{LlamaIndex Prometheus-2 Cookbook}} with GPT-4 and Gemini-2.5 Pro, as proposed by \citet{kim2024prometheus}, which provides a rubric-based score ranging from 1 to 5. For CRUD tasks, we first call the \texttt{read()} function to verify whether the task was executed correctly, and then apply rubric-based scoring to the \texttt{read()} output. In addition to automated evaluation, we conduct human evaluation focusing on two aspects: (a) whether the agent successfully completed the task, and (b) experts are required to complete the task. A separate set of experts then assess the correctness of these human-executed tasks. Scores are averaged across three experts serving as annotators.
\\
\noindent  For the evaluation of SFT and DPO, the trained model generates planning or action steps, and the LangChain framework is used to execute the tasks. Evaluation is performed using Prometheus-2 with Gemini-2.5 Pro, consistent with the evaluation methodology applied to the CAI systems.


\renewcommand{\arraystretch}{1.1} 

\begin{table*}[t]
\scriptsize
\centering
\resizebox{\linewidth}{!}{%
\begin{tabularx}{\linewidth}{l*{8}{>{\centering\arraybackslash}X}}
\hline
\textbf{Model} 
& \multicolumn{4}{c}{\textbf{GPT-4 Evaluator}} 
& \multicolumn{4}{c}{\textbf{Gemini-2.5 Pro Evaluator}} 
\\
\cline{2-9}
& \textbf{w/o Planning} & \textbf{CoT \cite{wei2022chain}} 
& \textbf{ReAct \cite{yao2022react}} & \cellcolor{gray!20}\textbf{w/ Gold Planning} 
& \textbf{w/o Planning} & \textbf{CoT \cite{wei2022chain}} 
& \textbf{ReAct \cite{yao2022react}} & \cellcolor{gray!20}\textbf{w/ Gold Planning} 
\\
\hline

\rowcolor{cyan!20}
\multicolumn{9}{c}{\textbf{LangChain Framework \cite{LangChain_2024}}} \\
\hline
\rowcolor{cyan!10} 
GPT-4o & 0.29 & 0.27 & 0.32 & \cellcolor{gray!20}0.43 
        & 0.27 & 0.28 & 0.29 & \cellcolor{gray!20}0.44 \\
\rowcolor{cyan!10} 
Claude-3.5-Sonnet & 0.31 & 0.27 & 0.28 & \cellcolor{gray!20}0.38 
        & 0.32 & 0.30 & 0.30 & \cellcolor{gray!20}0.41 \\
\rowcolor{cyan!10} 
o1-mini & 0.31 & 0.28 & 0.35 & \cellcolor{gray!20}0.51
        & 0.28 & 0.27 & 0.39 & \cellcolor{gray!20}0.47 \\
\rowcolor{cyan!10} 
Llama-3.1-8B & 0.04 & 0.06 & 0.14 & \cellcolor{gray!20}0.20 
        & 0.03 & 0.04 & 0.13 & \cellcolor{gray!20}0.21 \\
\rowcolor{cyan!10} 
Llama-3.3-70B & 0.23 & 0.22 & 0.21 & \cellcolor{gray!20}0.40
        & 0.24 & 0.23 & 0.21 & \cellcolor{gray!20}0.36 \\
\hline

\rowcolor{green!20}
\multicolumn{9}{c}{\textbf{DSPy \cite{khattab2024DSPy}}} \\
\hline
\rowcolor{green!10} 
GPT-4o & 0.19 & 0.32 & 0.34 & \cellcolor{gray!20}0.50
        & 0.25 & 0.26 & 0.29 & \cellcolor{gray!20}0.47 \\
\rowcolor{green!10} 
Claude-3.5-Sonnet & 0.19 & 0.24 & 0.30 & \cellcolor{gray!20}0.50
        & 0.21 & 0.29 & 0.29 & \cellcolor{gray!20}0.44 \\
\rowcolor{green!10} 
o1-mini & 0.29 & 0.33 & 0.38 & \cellcolor{gray!20}0.62 
        & 0.27 & 0.32 & 0.41 & \cellcolor{gray!20}0.63 \\
\rowcolor{green!10} 
Llama-3.1-8B & 0.10 & 0.14 & 0.16 & \cellcolor{gray!20}0.34 
        & 0.07 & 0.15 & 0.15 & \cellcolor{gray!20}0.34 \\
\rowcolor{green!10} 
Llama-3.3-70B & 0.20 & 0.27 & 0.30 & \cellcolor{gray!20}0.47 
        & 0.24 & 0.25 & 0.28 & \cellcolor{gray!20}0.48\\
\hline

\end{tabularx}
}
\caption{\textbf{\name \space Evaluation}: Comparison of performance across agents using different models and planning strategies with LangChain and DSPy frameworks, evaluated by GPT-4 and Gemini 2.5 Pro on 500 samples.}
\label{tab:model_planning_comparison}
\end{table*}


\begin{figure}[htbp]
    \centering
    \begin{subfigure}[b]{0.45\columnwidth}
        \centering
        \includegraphics[width=\textwidth]{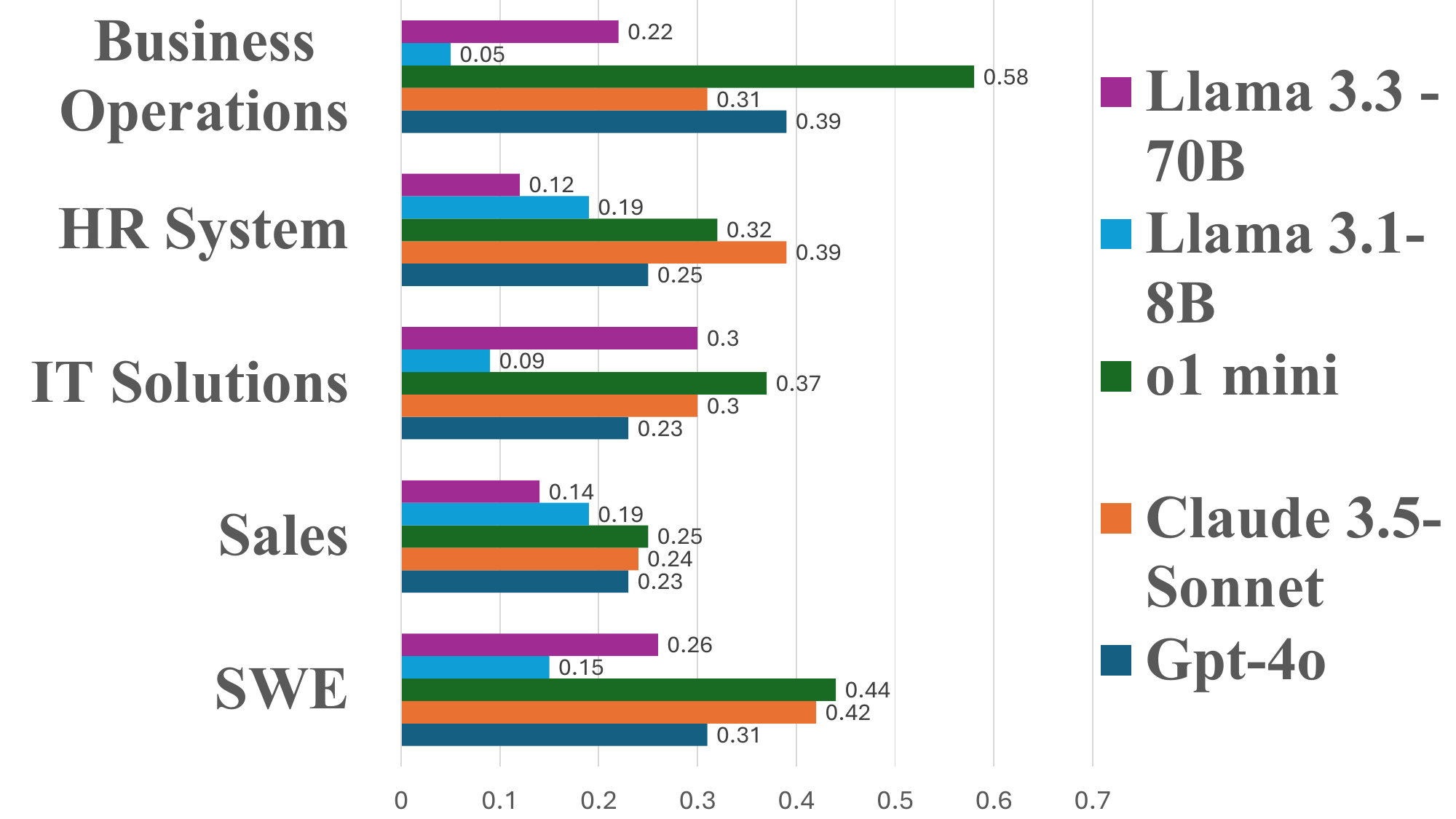}
        \caption{Performance of LangChain ReACT across different Domains}
        \label{fig:langchain_react}
    \end{subfigure}
    \hfill
    \begin{subfigure}[b]{0.45\columnwidth}
        \centering
        \includegraphics[width=\textwidth]{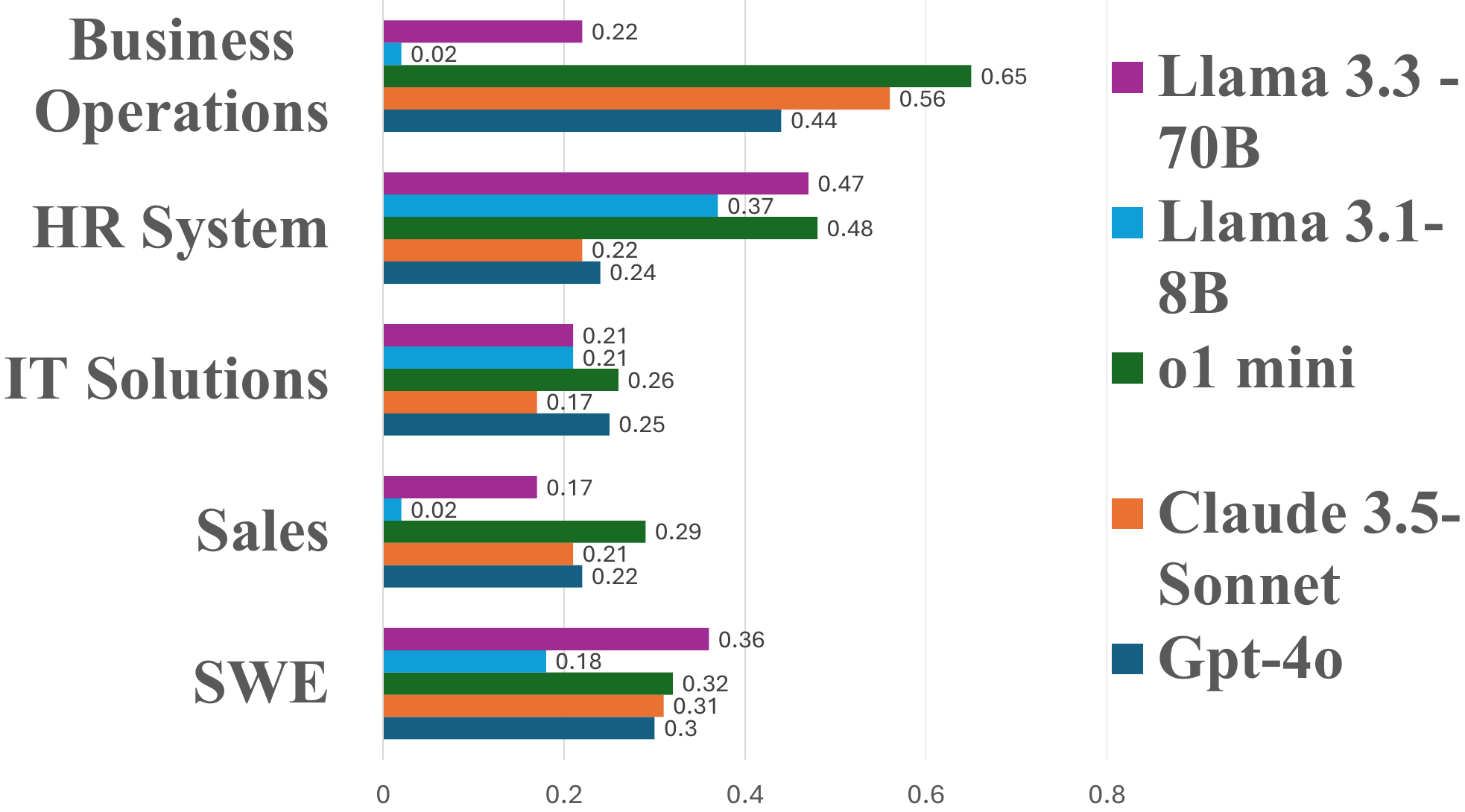}
        \caption{Performance of DSPy ReACT across different Domains}
        \label{fig:DSPy_react}
    \end{subfigure}
    \caption{Comparison of different models using ReAct planning: Performance across different domains of \name.}
    \label{fig:task_domain_performance}
\end{figure}

\vspace{-0.2cm}
\section{Results and Analysis}
\vspace{-0.2cm}
In this section, we evaluate our benchmark, \name, using five LLM agents built with state-of-the-art reasoning models: GPT-4o, Claude 3.5 Sonnet, Llama 3.1 8B, Llama 3.3 70B, and O1-mini. The agents are tested under different planning strategies implemented via LangChain and DSPy. We further report results from human evaluation, assessing both the correctness of agent responses and the successful execution of tasks. In addition, we present results from a model trained on \name, and provide an in-depth analysis of the evaluation outcomes for CAI systems.


\vspace{-0.2cm}
\subsection{Evaluation on Enterprise Search Tasks}
\textbf{Compound AI System Evaluation} Table~\ref{tab:model_planning_comparison} presents the evaluation of our benchmark across various models, planning strategies, and frameworks, scored using Prometheus-2 with GPT-4. ReAct-based planning outperforms both no-planning and CoT approaches across both frameworks. Among the models, O1-mini achieves the best performance, as expected given its advanced reasoning capabilities. The open-source Llama models show a significant performance drop compared to the higher-performing models, highlighting the need to improve their planning abilities. Notably, gold planning yields the highest accuracies, with approximately 40\% to 50\% improvements over ReAct. This substantial difference underscores the necessity for more sophisticated agents and frameworks capable of handling complex planning tasks in enterprise settings, which require coordination across multiple sources, tools, and function calls to successfully complete the final task. We also report performance across all domains using ReAct planning in figure \ref{fig:task_domain_performance}. Additionally, human evaluation was conducted on the agent built with O1-mini using ReAct planning within the LangChain framework, demonstrating an accuracy of 31\%.
\\
\noindent To further evaluate the performance of current LLM agents, we conducted a human CAI (humans acting as LLM agents) study to assess task execution. The accuracy achieved was 70\%, highlighting the gap between human performance and that of LLM agents in the enterprise setting. While human agents achieved higher accuracy, this came at the cost of significantly increased average completion time—from {50 seconds} with agents to {8 minutes 30 seconds per task} with humans—revealing a clear trade-off between precision and efficiency. These findings suggest that there is room to improve planning strategies in current LLM agents to achieve precision levels comparable to humans while maintaining significantly faster execution times.
\\
\noindent 
\textbf{Trained Model Evaluation} We conducted an additional experiment by training the Qwen3-8B model on data generated through our task generation pipeline. The model was fine-tuned using both supervised fine-tuning (SFT) and direct preference optimization (DPO) to predict planning or execution steps based on the task and available tools, with task execution carried out through the LangChain framework alongside GPT-4o. As shown in Table~\ref{tab:small_comparison}, Qwen3-8B achieved 27\% accuracy with SFT and 29\% with SFT+DPO on 1.2k samples, closely approaching GPT-4o with CoT. These results highlight the effectiveness of our benchmark and task generation pipeline, showing that even with limited training data, small models can achieve competitive performance with, and in some cases surpass, larger LLMs such as GPT-4o. This provides a proof of concept that for domain-specific tasks, small language models (SLMs) trained with high-quality data can outperform general-purpose LLMs.

\begin{table}[t]
\centering
\scriptsize
\begin{tabular}{lccc}
\toprule
\textbf{Model} & \textbf{GPT-4o w/ CoT} & \textbf{Qwen3-8B (SFT)} & {\textbf{Qwen3-8B (SFT+DPO)}}  \\
\midrule
Score & 0.27 & 0.27 & \textbf{0.29}  \\
\bottomrule
\end{tabular}
\caption{\small Performance comparison across GPT-4o w/ CoT and Qwen3-8B models using the LangChain framework for task execution on 200 samples. DPO results are reported with 1.2k preference pairs.}
\label{tab:small_comparison}
\end{table}

\vspace{-0.2cm}
\subsection{In-Depth Analysis}
\label{sec:eval_analysis}
We conduct an error analysis of the \texttt{O1-mini ReAct} agent implemented with LangChain. The evaluation was performed on 100 \name tasks, uniformly distributed across domains. The agent achieved an accuracy of 31\%, with the remaining cases classified as failures. Below, we outline the key failure modes identified through human evaluation.
\vspace{-0.2cm}
\begin{itemize}[noitemsep, left=0pt]
    \item \textit{Wrong Tool Selection / Wrong App Selection (18):} These errors arise from the complexity of tasks requiring multiple tool calls, as well as limitations in the model architectures used by LLM agents. We observed that models such as o1-mini perform slightly better in this regard compared to GPT, Claude, and other open-source models. Domain-wise performance, presented in Table \ref{tab:eval_f1} and Table \ref{tab:eval_gpt4} in the Appendix, shows that GPT performs well in HR and IT tasks, Claude excels in coding tasks, and o1-mini outperforms others in several non-technical domains. Performance in this area could be improved by incorporating continual learning, which would enhance the agent’s ability to understand the environment and make more accurate tool selections.

    \item \textit{Search-based Answer Hallucination (8):} The agent sometimes relies on prior knowledge instead of the retrieved context, leading to hallucinations such as fabricated policy names, incorrect dates, or non-existent entities, thereby compromising factual accuracy. This limitation could be mitigated through improved agent memory management.

    \item \textit{Context Retrieval (2):} The agent sometimes retrieves incomplete or irrelevant enterprise context due to weak query formulation or mismatches between the retrieval index and task intent, which leads to incorrect responses. Improving retriever performance requires going beyond similarity matching.

    \item \textit{Task Decomposition (20):} These errors often arise from the complexity of the tasks and the agents’ limited understanding of the sandbox environment. Performance in this area could be improved by employing a trained LLM agent rather than relying solely on general knowledge and few-shot examples.

    \item \textit{Partial Factual Coverage (14):} Some answers align with task goals but omit critical structured details (e.g., employee IDs, policy names, dates), reducing reliability and highlighting the need for precision in enterprise settings. Performance can be improved by using constrained decoding or function-calling approaches, which ensure that all required structured fields are consistently produced.

    \item \textit{Final Step Execution (7):} Even with correct subgoals, the final synthesis step may miscombine results, leading to incorrect answers and exposing gaps in temporal or logical consistency. Performance in this area could be improved by incorporating step validation or structured reasoning mechanisms to ensure accurate integration of intermediate outputs.
\end{itemize}

\noindent Our findings highlight that enterprise agents require tighter coupling between planning, retrieval, and grounding mechanisms, along with robustness against hallucinations and tool invocation errors. These insights aim to support the development of next-generation agentic systems that meet the strict accuracy demands of enterprise environments.

\vspace{-0.2cm}
\section{Conclusion}
\label{sec: conclusion}
\vspace{-0.2cm}
In this paper, we highlight the importance of Compound AI Systems in enterprise settings and the need for a benchmark to evaluate their performance. To address this, we introduce \name, a novel benchmark designed to assess CAI systems on complex enterprise tasks. Our experiments show that even state-of-the-art agents face significant challenges with these tasks. To create an evaluation environment, we develop an enterprise sandbox and a task framework, enabling the construction of comprehensive benchmark with minimal input.

\section*{Limitations}

The limitations of our work are as follows: 1) Our enterprise data generation process requires an initial set of real enterprise data, which can be costly to obtain. Relying solely on synthetic data may affect the realism of generated tasks. 2) Human experts are needed to verify intermediate steps during task generation, adding to the complexity and cost. 3) While we achieve high accuracy in enterprise task generation, some errors remain, suggesting areas for future improvement. 4) The evaluation of our benchmark relies on the current capabilities of reasoning models, which are likely to improve over time. 5) Our experiments did not involve large-scale data generation with terabytes of data, which would better represent real-world enterprise-scale scenarios.

\section*{Acknowledgement}
We thank the members of the AI Lab at Fujitsu Research for their valuable feedback on this work. We are also deeply grateful to the anonymous ARR reviewers, the meta-reviewer, and the ACL program chairs for their thoughtful comments and suggestions, which significantly improved the paper.


\bibliography{custom}

\clearpage

\appendix


\section{Appendix}
\label{sec:appendix}

\noindent In this section, we present additional results and analyses that could not be included in the main paper due to space constraints. It also includes visual illustrations of the sandbox environment for \name, and LLM prompts used for benchmark creation and baseline execution. Specifically, this appendix contains the following:

\begin{itemize}[nosep]
    \item \hyperref[sec:add_res]{Algorithms, Additional Results, and Details}
        \item \hyperref[sec:abl_study]{Ablation Study}
    \item \hyperref[sec:exp_study]{Expert Study Details}
    \item \hyperref[sec:sandbox]{Details of simulating the Enterprise Sandbox}

    
    \item \hyperref[sec:prompts]{LLM Prompts}
\end{itemize}

\subsection{Additional Results, Algorithm, and Details}
\label{sec:add_res}

\textbf{Algorithm} To generate tasks tailored to individual enterprise employees, we design a pipeline that dynamically incorporates employee context, role-specific goals, and relevant enterprise entities. The process begins by retrieving contextual information based on the employee’s ID and domain of interest, followed by the selection of a suitable goal template. This goal is expanded into subgoals using contextual and entity-aware reasoning. Templates are then populated to construct a task instance, which is iteratively refined and validated using LLM capabilities. The full task generation procedure is detailed in Algorithm~\ref{alg:generate_emp_task}.
\\\\
\noindent \textbf{Additional Results} Table \ref{tab:eval_f1} shows the evaluation of \name \space using F1 score as the metric across five domains in our benchmark: SWE, Sales, HR, IT, and Business Development. This table allows us to observe the performance of tasks within each domain, which can guide future development of better agents tailored for enterprise settings through separate domain evaluations. Additionally, Table \ref{tab:eval_gpt4} presents the evaluation results using Prometheus-2 with GPT-4 across domains.
\\\\
\noindent \textbf{Tools Inventory} Table \ref{tab:app_tools} presents the collection of tools and functions developed for our \name \space sandbox environment to support the operation of the LLM Agents.
\\\\
\noindent \textbf{Post Training Data format} We conducted SFT\footnote{\href{https://huggingface.co/docs/trl/en/sft_trainer\#expected-dataset-type-and-format}{SFT Trainer Data Format}} and DPO\footnote{\href{https://huggingface.co/docs/trl/en/dataset_formats\#preference}{DPO Data Format}} fine-tuning experiments using the standard dataset formats, illustrated in Listing~\ref{lst:sft} and Listing~\ref{lst:dpo}, respectively.

\lstdefinelanguage{json}{
  basicstyle=\ttfamily\small,
  numbers=left,
  numberstyle=\tiny,
  stepnumber=1,
  numbersep=5pt,
  showstringspaces=false,
  breaklines=true,
  frame=single,
  literate=
   *{0}{{{\color{black}0}}}{1}
    {1}{{{\color{black}1}}}{1}
    {2}{{{\color{black}2}}}{1}
    {3}{{{\color{black}3}}}{1}
    {4}{{{\color{black}4}}}{1}
    {5}{{{\color{black}5}}}{1}
    {6}{{{\color{black}6}}}{1}
    {7}{{{\color{black}7}}}{1}
    {8}{{{\color{black}8}}}{1}
    {9}{{{\color{black}9}}}{1}
}

\begin{lstlisting}[language=json, caption={SFT data format used in training}, label={lst:sft}]
{
  "messages": [
    {"role": "system", "content": "You are a helpful assistant"},
    {"role": "user", "content": "What color is the sky?"},
    {"role": "assistant", "content": "It is blue."}
  ]
}
\end{lstlisting}

\begin{lstlisting}[language=json, caption={DPO data format used in training}, label={lst:dpo}]
{
  "prompt": [
    {"role": "user", "content": "What color is the sky?"}
  ],
  "chosen": [
    {"role": "assistant", "content": "It is blue."}
  ],
  "rejected": [
    {"role": "assistant", "content": "It is green."}
  ]
}
\end{lstlisting}


\begin{table}[h!]
\centering
\small
\resizebox{\columnwidth}{!}{%
\begin{tabular}{lcccc}
\toprule
\textbf{Benchmark} & \textbf{Coverage} & \textbf{Complexity} & \textbf{Diversity} & \textbf{Expert Validation} \\
\midrule
SWEBench \cite{jimenezswe} & 2  & 0.5    & 1 & \xmark \\
WorkArena  \cite{drouin2024workarena}     & 7  & 0.86 & 6 & \xmark \\
WorkBench \cite{stylesworkbench} & 5 & 0 & 5 & \xmark \\
AgentBench \cite{liuagentbench} & 8  & 0   & 8 & \xmark \\
$\tau$-bench \cite{yaotau} & 3 & 0.67 & 8 & \xmark\\
CRMArena \cite{huang-etal-2025-crmarena} & 16 & 1.3 & 9 & \cmark\\
\hline
EnterpriseBench (Ours) & 17 & 1.2 & 17 & \cmark\\
\bottomrule
\end{tabular}%
}
\caption{\small {Comparison of benchmarks in terms of: coverage (\# objects that mirror core components in the simulated environment; ER diagram nodes), environment complexity (\# dependencies/object; average connections in ER diagram), and diversity (classification of tasks spread across domains).}}
\label{tab:benchmark_comparison}
\end{table}


\begin{algorithm}
\scriptsize  
\caption{Generate Employee-Specific Task}
\label{alg:generate_emp_task}
\begin{algorithmic}[1]
\Function{Generate}{emp\_id, persona, config, tools, task\_domain, task\_category}
    \State context $\gets$ \Call{GetContext}{emp\_id, config["source\_paths"], task\_domain, task\_category}
    \State goal $\gets$ \Call{ChooseGoal}{config["goal\_templates"], task\_domain, task\_category}
    \State entities $\gets$ \Call{EntityExtraction}{tools, context, goal}
    \State subGoals $\gets$ \Call{GetSubgoal}{ goal, entities, context}
    \State templates $\gets$ \Call{GetTemplate}{subGoals, entities, context, persona}
    \State task $\gets$ \Call{GetTask}{goal, subGoals, entities, templates, context, persona}
    
    \For{$i = 1$ to max\_iter}
        \If{\Call{Validate}{task}} \Return task \EndIf
        \State task $\gets$ \Call{Rephrase}{task}
    \EndFor
    \State \Return task
\EndFunction
\end{algorithmic}
\end{algorithm}
\begin{table*}[ht]
\centering
\scriptsize
\setlength{\tabcolsep}{3pt} 
\begin{adjustbox}{max width=\textwidth}
\begin{tabular}{|l|ccccc|ccccc|}
\hline
\textbf{Department} & \multicolumn{5}{c|}{\textbf{LangChain}} & \multicolumn{5}{c|}{\textbf{DSPy}} \\
 & GPT-4o & Claude-3.5 & o1-mini & LLaMA-3.1-8B & LLaMA-3.1-70B & GPT-4o & Claude-3.5 & o1-mini & LLaMA-3.1-8B & LLaMA-3.1-70B \\
\hline

\rowcolor{gray!20}
\multicolumn{11}{|c|}{\textbf{w/o Planning}} \\
SWE           & 0.29 & 0.24 & 0.26 & 0.03 & 0.29 & 0.24 & 0.20 & 0.26 & 0.07 & 0.28 \\
Sales         & 0.20 & 0.26 & 0.22 & 0.03 & 0.25 & 0.25 & 0.21 & 0.25 & 0.07 & 0.29 \\
HR            & 0.42 & 0.49 & 0.46 & 0.03 & 0.12 & 0.33 & 0.20 & 0.29 & 0.12 & 0.33 \\
IT            & 0.32 & 0.31 & 0.32 & 0.03 & 0.31 & 0.32 & 0.24 & 0.28 & 0.13 & 0.32 \\
Business Ops  & 0.26 & 0.37 & 0.36 & 0.03 & 0.32 & 0.25 & 0.36 & 0.27 & 0.03 & 0.15 \\
\hline

\rowcolor{gray!20}
\multicolumn{11}{|c|}{\textbf{CoT \cite{wei2022chain}}} \\
SWE           & 0.27 & 0.23 & 0.23 & 0.02 & 0.22 & 0.27 & 0.21 & 0.29 & 0.16 & 0.29 \\
Sales         & 0.28 & 0.23 & 0.27 & 0.02 & 0.25 & 0.27 & 0.21 & 0.30 & 0.06 & 0.06 \\
HR            & 0.47 & 0.34 & 0.44 & 0.03 & 0.30 & 0.37 & 0.25 & 0.35 & 0.14 & 0.35 \\
IT            & 0.35 & 0.28 & 0.30 & 0.03 & 0.25 & 0.34 & 0.27 & 0.31 & 0.21 & 0.31 \\
Business Ops  & 0.33 & 0.37 & 0.31 & 0.03 & 0.35 & 0.35 & 0.37 & 0.30 & 0.03 & 0.33 \\
\hline

\rowcolor{gray!20}
\multicolumn{11}{|c|}{\textbf{ReAct \cite{yao2022react}}} \\
SWE           & 0.27 & 0.23 & 0.24 & 0.02 & 0.22 & 0.25 & 0.20 & 0.31 & 0.11 & 0.31 \\
Sales         & 0.29 & 0.23 & 0.29 & 0.12 & 0.25 & 0.12 & 0.21 & 0.32 & 0.08 & 0.17 \\
HR            & 0.48 & 0.29 & 0.43 & 0.14 & 0.31 & 0.47 & 0.30 & 0.38 & 0.03 & 0.40 \\
IT            & 0.37 & 0.28 & 0.28 & 0.13 & 0.22 & 0.24 & 0.27 & 0.33 & 0.19 & 0.30 \\
Business Ops  & 0.31 & 0.39 & 0.42 & 0.13 & 0.19 & 0.49 & 0.35 & 0.37 & 0.13 & 0.36 \\
\hline

\rowcolor{gray!20}
\multicolumn{11}{|c|}{\textbf{w/ Gold Planning}} \\
SWE           & 0.36 & 0.40 & 0.35 & 0.14 & 0.42 & 0.44 & 0.38 & 0.48 & 0.37 & 0.42 \\
Sales         & 0.26 & 0.27 & 0.28 & 0.13 & 0.27 & 0.32 & 0.25 & 0.44 & 0.13 & 0.28 \\
HR            & 0.48 & 0.33 & 0.49 & 0.14 & 0.48 & 0.57 & 0.41 & 0.55 & 0.40 & 0.53 \\
IT            & 0.37 & 0.40 & 0.38 & 0.14 & 0.33 & 0.43 & 0.43 & 0.49 & 0.31 & 0.39 \\
Business Ops  & 0.45 & 0.56 & 0.61 & 0.24 & 0.34 & 0.51 & 0.47 & 0.51 & 0.33 & 0.47 \\
\hline
\end{tabular}
\end{adjustbox}
\caption{\textbf{\name \space Evaluation}: Domain-wise performance comparison using F1 score.}
\label{tab:eval_f1}
\end{table*}


\begin{table*}[ht]
\centering
\scriptsize
\setlength{\tabcolsep}{3pt} 
\begin{adjustbox}{max width=\textwidth}
\begin{tabular}{|l|ccccc|ccccc|}
\hline
\textbf{Department} & \multicolumn{5}{c|}{\textbf{LangChain}} & \multicolumn{5}{c|}{\textbf{DSPy}} \\
 & GPT-4o & Claude-3.5 & o1-mini & LLaMA-3.1-8B & LLaMA-3.1-70B & GPT-4o & Claude-3.5 & o1-mini & LLaMA-3.1-8B & LLaMA-3.1-70B \\
\hline

\rowcolor{gray!20}
\multicolumn{11}{|c|}{\textbf{w/o Planning}} \\
SWE           & 0.30 & 0.19 & 0.30 & 0.02 & 0.21 & 0.17 & 0.21 & 0.33 & 0.13 & 0.24 \\
Sales         & 0.16 & 0.20 & 0.17 & 0.06 & 0.14 & 0.16 & 0.15 & 0.21 & 0.11 & 0.20 \\
HR            & 0.36 & 0.62 & 0.39 & 0.05 & 0.47 & 0.14 & 0.13 & 0.18 & 0.12 & 0.18 \\
IT            & 0.19 & 0.23 & 0.33 & 0.07 & 0.19 & 0.15 & 0.11 & 0.35 & 0.10 & 0.21 \\
Business Ops  & 0.33 & 0.30 & 0.33 & 0.00 & 0.15 & 0.33 & 0.37 & 0.36 & 0.05 & 0.18 \\
\hline

\rowcolor{gray!20}
\multicolumn{11}{|c|}{\textbf{CoT \cite{wei2022chain}}} \\
SWE           & 0.33 & 0.23 & 0.27 & 0.09 & 0.20 & 0.30 & 0.25 & 0.30 & 0.12 & 0.30 \\
Sales         & 0.16 & 0.12 & 0.20 & 0.06 & 0.21 & 0.16 & 0.08 & 0.15 & 0.02 & 0.09 \\
HR            & 0.44 & 0.40 & 0.40 & 0.05 & 0.40 & 0.45 & 0.30 & 0.48 & 0.15 & 0.48 \\
IT            & 0.31 & 0.15 & 0.16 & 0.04 & 0.19 & 0.28 & 0.18 & 0.22 & 0.14 & 0.30 \\
Business Ops  & 0.20 & 0.20 & 0.38 & 0.05 & 0.10 & 0.40 & 0.37 & 0.50 & 0.05 & 0.20 \\
\hline

\rowcolor{gray!20}
\multicolumn{11}{|c|}{\textbf{ReAct \cite{yao2022react}}} \\
SWE           & 0.34 & 0.27 & 0.31 & 0.15 & 0.26 & 0.28 & 0.22 & 0.32 & 0.19 & 0.37 \\
Sales         & 0.14 & 0.16 & 0.26 & 0.20 & 0.15 & 0.13 & 0.09 & 0.20 & 0.03 & 0.17 \\
HR            & 0.44 & 0.39 & 0.32 & 0.09 & 0.31 & 0.52 & 0.46 & 0.48 & 0.37 & 0.54 \\
IT            & 0.34 & 0.19 & 0.27 & 0.19 & 0.13 & 0.22 & 0.17 & 0.26 & 0.21 & 0.21 \\
Business Ops  & 0.20 & 0.41 & 0.58 & 0.05 & 0.23 & 0.54 & 0.57 & 0.65 & 0.03 & 0.23 \\
\hline

\rowcolor{gray!20}
\multicolumn{11}{|c|}{\textbf{w/ Gold Planning}} \\
SWE           & 0.50 & 0.60 & 0.63 & 0.13 & 0.59 & 0.65 & 0.62 & 0.80 & 0.38 & 0.63 \\
Sales         & 0.16 & 0.21 & 0.21 & 0.23 & 0.13 & 0.27 & 0.15 & 0.20 & 0.14 & 0.16 \\
HR            & 0.54 & 0.13 & 0.57 & 0.19 & 0.56 & 0.62 & 0.68 & 0.81 & 0.44 & 0.67 \\
IT            & 0.46 & 0.47 & 0.53 & 0.12 & 0.42 & 0.58 & 0.51 & 0.68 & 0.50 & 0.49 \\
Business Ops  & 0.40 & 0.50 & 0.60 & 0.30 & 0.33 & 0.40 & 0.56 & 0.62 & 0.25 & 0.40 \\
\hline
\end{tabular}
\end{adjustbox}
\caption{\textbf{\name \space Evaluation}: Domain-wise performance comparison with Prometheus-2 using GPT-4 score.}
\label{tab:eval_gpt4}
\end{table*}

\begin{table*}[t]
\centering
\footnotesize
\caption{List of Apps, Tools, and their Descriptions}
\label{tab:app_tools}
\renewcommand{\arraystretch}{1.2}
\begin{tabular}{>{\centering\arraybackslash}m{1.8cm}
                >{\centering\arraybackslash}m{4cm}
                >{\centering\arraybackslash}m{8.0cm}}
\toprule
\textbf{App} & \textbf{Tool} & \textbf{Description} \\
\midrule
HR System & \texttt{employee\_data\_read} & Reads or Fetches the employee record based on emp\_id, name, email or semantic query (vector DB). \\
HR System & \texttt{employee\_data\_create} & Creates a new employee record. \\
HR System & \texttt{employee\_data\_update} & Updates an existing employee record. \\
HR System & \texttt{employee\_data\_delete} & Deletes an employee record (sets is\_valid to False). \\
Enterprise Mail System & \texttt{enterprise\_mail} \texttt{\_system\_read} & Reads email data based on sender's email, recipient's email, optional thread ID, and optional email ID or semantic query (vector DB). Retrieves user-specific emails, emails in a thread, or a specific conversation while enforcing access control. \\
Enterprise Mail System & \texttt{enterprise\_mail} \texttt{\_system\_create} & Creates a new email conversation. Generates unique email and thread IDs, validates participants, and stores the email in the database. \\
Enterprise Mail System & \texttt{enterprise\_mail} \texttt{\_system\_update} & Updates an existing email conversation by modifying the subject, body, importance, category, signature, or confidentiality notice. Ensures user authorization before making changes. \\
Enterprise Mail System & \texttt{enterprise\_mail} \texttt{\_system\_delete} & Deletes an email conversation based on user ID, thread ID, and email ID. Ensures the user is a participant and has permission before removing the email. \\
Chats & \texttt{conversations\_create} & Creates a new conversation entry in a particular software engineer team between emp1 and emp2 if the conversation ID does not already exist. \\
Chats & \texttt{conversations\_read} & Reads and displays a conversation between emp1 and emp2 based on their emp\_id's, conversation\_id or semantic query (vector DB). Only authorized employees either emp1 or emp2 can access the conversation. \\
Chats & \texttt{conversations\_update} & Updates an existing conversation entry if the conversation exists between emp1 and emp2 of software engineer team. \\
Chats & \texttt{conversations\_delete} & Deletes an existing conversation entry if the conversation exists between emp1 and emp2 of software engineer team, the employee is authorized to delete it, and the repository is valid. \\
Workspace & \texttt{github\_read} & Reads and displays the GitHub repository along with their issues based on specific path, repo\_name or semantic query (vector DB) or fetches all repositories accessible to an employee. Ensures access control before retrieving data. \\
Workspace & \texttt{github\_create} & Creates a new GitHub code entry. Ensures the path is unique, the employee is valid, and generates a hash for content integrity. \\
Workspace & \texttt{github\_update} & Updates an existing GitHub code entry. Ensures the path exists, the employee has sufficient access rights, and updates content with a new hash if modified. \\
Workspace & \texttt{github\_delete} & Deletes a GitHub code entry. Ensures the employee has the appropriate access rights and removes the code entry if permitted. \\
Sales & \texttt{products\_create} & Creates a new product entry if the product ID doesn't already exist, and the employee has the required access level. \\
Sales & \texttt{products\_read} & Reads and displays product details based on product\_id or semantic query (vector DB) if the employee has the required access level and the product ID exists. \\
Sales & \texttt{products\_update} & Updates an existing product entry if the product ID exists, and the employee has the required access level. Displays the product details before and after the update. \\
Sales & \texttt{products\_delete} & Deletes an existing product entry if the product ID exists, and the employee has the required access level. Confirms deletion by attempting to read the product after deletion. \\
Sales & \texttt{product\_sentiment\_create} & Creates a new product sentiment entry if the product ID, customer ID, and employee have the necessary access rights. \\
Sales & \texttt{product\_sentiment\_read} & Reads and displays the product sentiment data based on product ID, customer ID or semantic query (vector DB) and employee access rights. \\
Sales & \texttt{product\_sentiment\_update} & Updates an existing product sentiment entry if the product ID, customer ID, and employee have the necessary access rights. \\
\bottomrule
\end{tabular}
\end{table*}

\begin{table*}[htbp]
\centering
\footnotesize
\label{tab:app_tools_b}
\resizebox{\textwidth}{!}{%
\begin{tabular}{>{\centering\arraybackslash}m{1.8cm}
                >{\centering\arraybackslash}m{4.0cm}
                >{\centering\arraybackslash}m{8.0cm}}
\toprule
\textbf{App} & \textbf{Tool} & \textbf{Description} \\
\midrule

Sales & \texttt{product\_sentiment\_delete} & Deletes a product sentiment entry if the product ID, customer ID, and employee have the necessary access rights. \\
Sales & \texttt{sales\_create} & Creates a new product entry if the product does not already exist and the employee has the required permissions. \\
Sales & \texttt{sales\_read} & Reads and displays product sales details based on the product ID, product\_name, customer\_id, customer\_name, data\_of\_purchase or semantic query (vector DB). \\
Sales & \texttt{sales\_update} & Updates an existing product sales entry if the product exists, the employee is authorized, and the required permissions are met. \\
Sales & \texttt{sales\_delete} & Deletes an existing product sales entry based on access rules and authorization. Ensures the product exists and the employee has the necessary permissions. \\
Sales &  \texttt{customer\_support\_chats} \texttt{\_create} & Creates a new support entry with employee ID, product ID, customer ID, text, and an optional interaction date. Validates inputs before appending data to the support log. \\
Sales & \texttt{customer\_support\_chats} \texttt{\_read} & Reads customer support chats between employee and customer based on employee ID, product ID, customer ID or semantic query (vector DB). Ensures employee validation and access control before retrieving records. \\
Sales & \texttt{customer\_support\_chats} \texttt{\_update} & Updates an existing support record with new text and an interaction date. Ensures the entry exists and the employee has necessary permissions before modifying data. \\
Sales & \texttt{customer\_support\_chats} \texttt{\_delete} & Deletes a support entry based on employee ID, product ID, and customer ID. Checks access control and validates inputs before removing the record. \\
IT Solutions & \texttt{it\_service\_management} \texttt{\_create\_issue} & Creates a new IT helpdesk issue if the employee has access and the issue ID doesn't exist. \\
IT Solutions & \texttt{it\_service\_management} \texttt{\_read\_issue} & Reads and displays a IT helpdesk issue that is raised by an employee with id raised\_by\_emp\_id and handled by IT department employee with employee ID emp\_id or semantic query (vector DB) if the employee has access. \\
IT Solutions & \texttt{it\_service\_management} \texttt{\_update\_issue} & Updates an existing helpdesk issue if the employee has access. \\
IT Solutions & \texttt{it\_service\_management} \texttt{\_delete\_issue} & Deletes a helpdesk issue if the employee has access. \\
Social Platform & \texttt{social\_platform\_create} & Creates a new post in the clubbed posts JSON. \\
Social Platform & \texttt{social\_platform\_read} & Reads a post from the clubbed posts JSON. \\
Social Platform & \texttt{social\_platform\_update} & Updates an existing post in the clubbed posts JSON. \\
Social Platform & \texttt{social\_platform\_delete} & Deletes a post from the clubbed posts JSON. \\
Internal Overflow & \texttt{overflow\_read} & Reads and displays an overflow post if the user has access. \\
Internal Overflow & \texttt{overflow\_create} & Creates a new overflow post if the user is a valid employee and the post ID is unique. \\
Internal Overflow & \texttt{overflow\_update} & Updates an existing overflow post if the user is the owner and a valid employee. \\
Internal Overflow & \texttt{overflow\_delete} & Deletes an overflow post if the user is the owner and a valid employee. \\
Enterprise Policy Documents & \texttt{document\_read} & Retrieves the relevant data from the policy document pds using ColPali \\
LLM Call & \texttt{llm\_call} & General purpose llm call \\
\bottomrule
\end{tabular}
}
\end{table*}

\begin{table*}[h!]
\centering
\footnotesize
\setlength{\tabcolsep}{10pt} 
\begin{tabular}{|c|P{12.5cm}|} 
\hline
\textbf{Domain} & \textbf{Goal Template} \\
\hline
Business Operation & Send an email about scheduling a meeting \\
\hline
Business Operation & Send an email about Performance Management \\
\hline
HR & Send an email requesting for leaves \\
\hline
HR & Update the Salary of employee [Employee ID] to \$150000 \\
\hline
Sales & Get the details of [Product ID or Product Name] (id, name, price, description) with the most reviews from customers. \\
\hline
Sales & Get the sentiment (positive/negative/neutral) from the customer's review content for [Product ID or Product Name]. \\
\hline
SWE & Update the metadata of a particular repository \\
\hline
SWE & List the names of all the GitHub repositories owned by [Employee ID] \\
\hline
IT & Get all my tickets on [date] with `high` priority, give ID and list them \\
\hline
IT & Send a message about new Asset Configurations \\
\hline
\end{tabular}
\caption{Domain experts curated task goal templates for the \name \space task curation, organized by domain.}
\label{tab:goal-templates}
\end{table*}

\noindent \textbf{Defining the Ground Truth}
Below, we summarize the step-by-step pipeline used to generate task-specific ground truth in a traceable and verifiable manner:

\begin{enumerate}[nosep]
    \item \textit{Retrieve Context:} Relevant data are fetched from pre-defined enterprise sources using employee ID, task domain, and task category.
    
    \item \textit{Extract Entities and Relations:} An LLM is employed to extract (i) entities (e.g., employee, GitHub repository name, issue ID) and (ii) relations (e.g., issues linked to a repository, metadata associated with a repository).
    
    \item \textit{Decompose Goal into Subtask Templates:} The primary task goal is decomposed into logical subtasks using LLMs, guided by domain-specific tools and the retrieved context.
    
    \item \textit{Fill Subtask Templates:} Extracted entities are inserted into subtask templates according to their semantic types.
    
    \item \textit{Ground Each Subtask:} Each subtask is linked to relevant contextual evidence (sentences or snippets) using the identified relations.
    
    \item \textit{Generate Final Task Ground Truth:} All subtasks and their grounded context are combined to form a complete, traceable task-level ground truth.
    
    \item \textit{Validation and Refinement:} The generated ground truth undergoes iterative refinement, after which human experts validate correctness and relevancy.
\end{enumerate}

\noindent This process follows a \textit{reverse task synthesis} paradigm: rather than generating answers to predefined questions, we start from the available context and a goal template. We then frame the most appropriate task whose subgoals and answers are already embedded in the context. This ensures that each task is grounded, domain-relevant, and verifiable.



\subsection{Ablation Study}
\label{sec:abl_study}

We perform ablation studies to analyze the effect of planning quality, task complexity, and access control on model performance.

\noindent \textbf{Gold Planning}
As shown in Table~\ref{tab:model_planning_comparison}, providing models with gold plans yields substantial improvements over other planning strategies, highlighting the critical role of accurate planning in complex task execution.

\noindent \textbf{Task Complexity}
We categorize tasks as \emph{easy} if they contain fewer than three subtasks, and \emph{hard} if they contain three or more. Table~\ref{tab:complexity_ablation} compares performance under the ReAct baseline versus our planning-enhanced approach.



\noindent The results show a clear drop in performance as task complexity increases (\textit{i.e.}, when the number of subtasks $\geq 3$). This ablation reveals that longer interaction trajectories increase failure rates, likely due to the models’ lack of prior knowledge of the sandbox environment and limited memory across steps. While planning improves robustness, it does not fully close this gap, highlighting the difficulty of long-horizon reasoning in unfamiliar environments.

\noindent \textbf{Access Control}
In this ablation, we remove the access control constraint on tasks classified as \emph{Unanswerable}. Without constraints, models attempt these tasks despite lacking the necessary permissions, leading to incorrect executions. These are counted as failures in our evaluation, confirming that access control mechanisms are essential to prevent spurious task completions.

\subsection{Expert Study Details}
\label{sec:exp_study}

We selected domain experts from various departments within the organization to assist in task evaluation, goal template design, and sandbox environment simulation. For the simulation and goal template creation, we engaged a group of 10 domain experts spanning the target domains. For task evaluation, we ensured relevant participation by circulating a Microsoft Form, requiring that respondents hold job titles aligned with roles defined in EnterpriseBench: Sales, Customer Support, Engineer, IT Support, and HR. Table~\ref{tab:profession_data} presents participant profiles involved in sandbox simulation and task validation.

\begin{figure*}[htbp]
    \centering
    \begin{subfigure}[b]{0.45\textwidth}
        \centering
        \includegraphics[height=6.0cm]{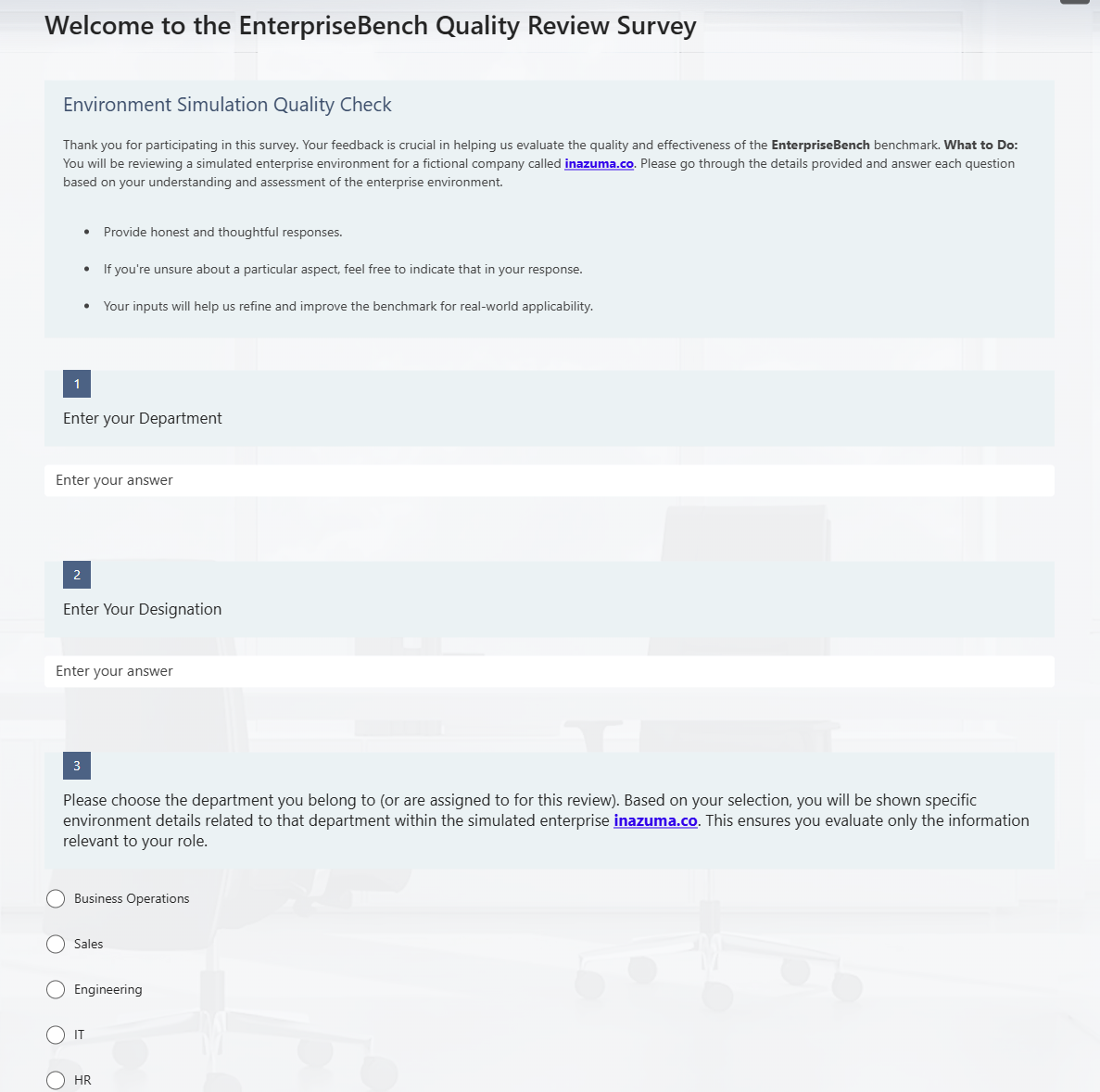}
        \caption{First page of the Microsoft Form used to collect information about domain experts, including their department and position.}
        \label{fig:img1}
    \end{subfigure}
    \hfill
    \begin{subfigure}[b]{0.60\textwidth}
        \centering
        \includegraphics[height=4.4cm]{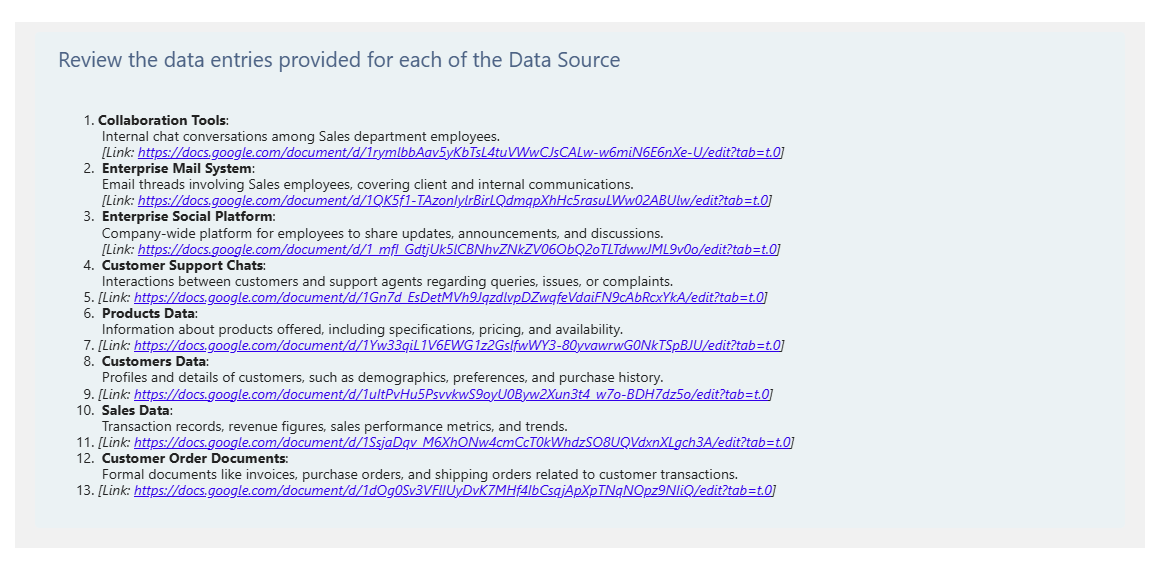}
        \caption{Next page of the form displaying simulated data details for the selected department. This example shows sales data from the enterprise.}
        \label{fig:img2}
    \end{subfigure}

    \vspace{0.5cm}

    \begin{subfigure}[b]{0.45\textwidth}
        \centering
        \includegraphics[height=6.0cm]{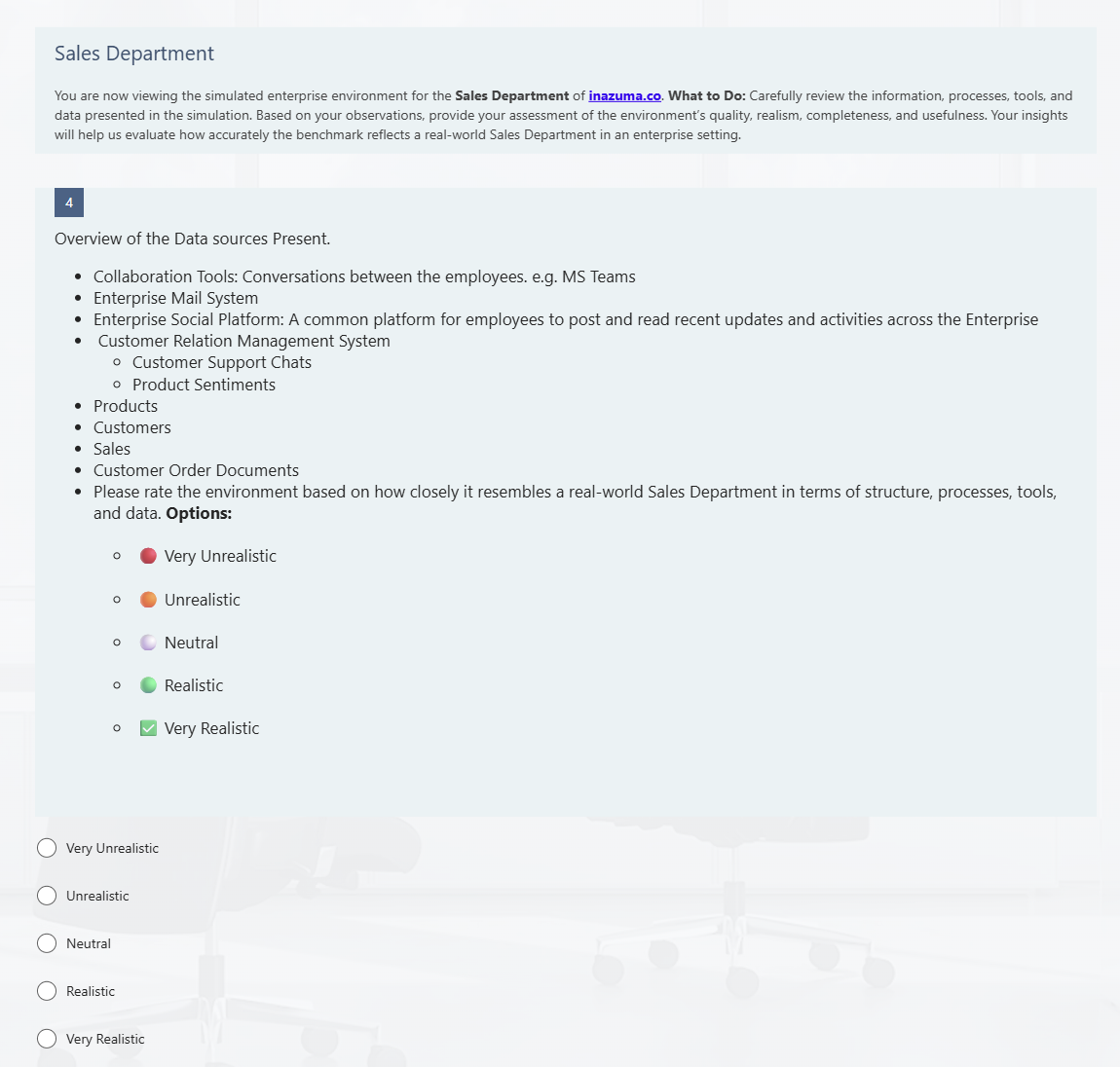}
        \caption{Users are asked to rate the realism of the simulated data for the selected department, choosing from options ranging from `Very Unrealistic’ to `Very Realistic.’ They also have to provide reasons when selecting `Unrealistic'.}
        \label{fig:img3}
    \end{subfigure}
    \hfill
    \begin{subfigure}[b]{0.45\textwidth}
        \centering
        \includegraphics[height=6.0cm]{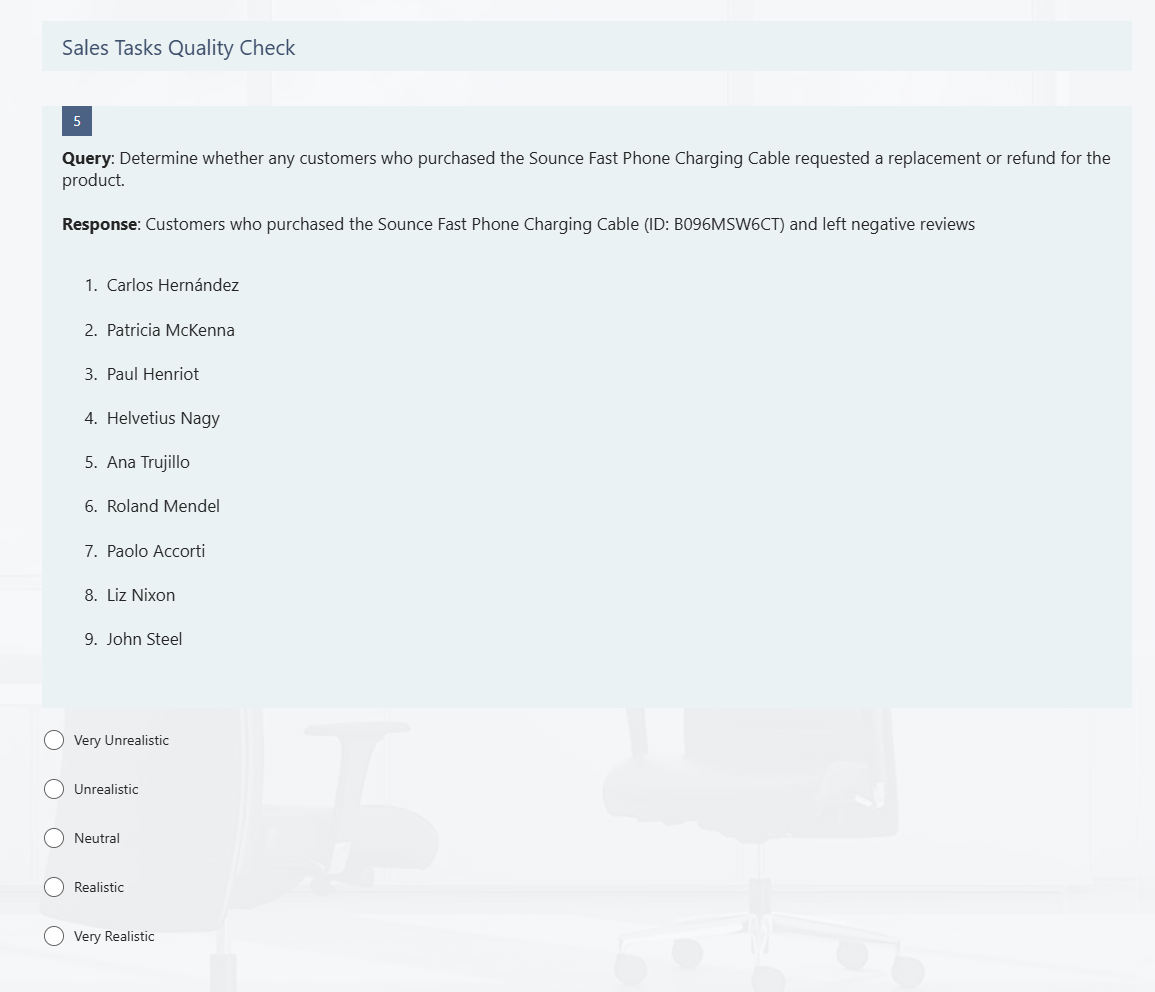}
        \caption{This page presents enterprise tasks for evaluation. Users rate each task’s realism from ‘Very Unrealistic’ to ‘Very Realistic,’ and provide reasons if they select `Neutral,’ `Unrealistic,’ or `Very Unrealistic'.}
        \label{fig:img4}
    \end{subfigure}

  \caption[\textbf{Domain Expert Validation for Realism of EnterpriseBench Tasks}]{\textbf{Domain Expert Validation in EnterpriseBench.} Domain experts from all benchmark domains evaluate the realism of the generated data and created tasks. This example shows screenshots of MS form for different steps a domain expert completes during the validation process.}

    \label{fig:cluster}
    \vspace{-0.4cm}
\end{figure*}

\noindent Details of the MS form (screenshots in figure \ref{fig:cluster} to validate the realism of enterprise data and tasks are provided below:

\begin{itemize}[nosep, left=0pt]
    \item \textit{Part-1:} (10 seconds) The participant (or expert) logs in by selecting their department and role (in figure \ref{fig:img1}).

    \item \textit{Part-2:} (15 minutes) After logging in, they are presented with instructions outlining the task: they are asked to assess the realism of the organizational environment-such as the employee flow chart and access control-and then evaluate the realism of the tasks, which are displayed on the following page (in figure \ref{fig:img2}, \ref{fig:img3}).

    \item \textit{Part-3:} (10 minutes/task) On the following pages, elements of the enterprise environment are shown, followed by role-specific tasks—such as emails, chats, and more—tailored to the participant’s selected department and role (see Figure \ref{fig:img4}). The user is required to perform these tasks in the sandbox and rate their realism.
\end{itemize}

\noindent The participants are asked to rate the realism of environment setup and tasks using below options:

\begin{enumerate}[nosep]

    \item \textit{Very Unrealistic:} The organizational structure and tasks seemed very artificial and didn't resemble how real organizations typically operate.

    \item \textit{Unrealistic:} While the organization and tasks included some familiar elements, many aspects lacked a convincing or realistic structure.

    \item \textit{Neutral:} The organization and tasks felt partially realistic, combining both plausible and implausible elements.

    \item \textit{Realistic:} The organization largely resembled a real-world setup, and the tasks reflected what an employee might typically ask, though there were minor inconsistencies.

    \item \textit{Very Realistic:} The organization appeared fully authentic, with a structure akin to real-world setups, and the tasks aligned well with those typically posed by employees.
    
\end{enumerate}

\subsection{Details of simulating the \name \space Sandbox}
\label{sec:sandbox}

\begin{figure*}[b]
        \centering
        \includegraphics[width=24cm, angle=90]{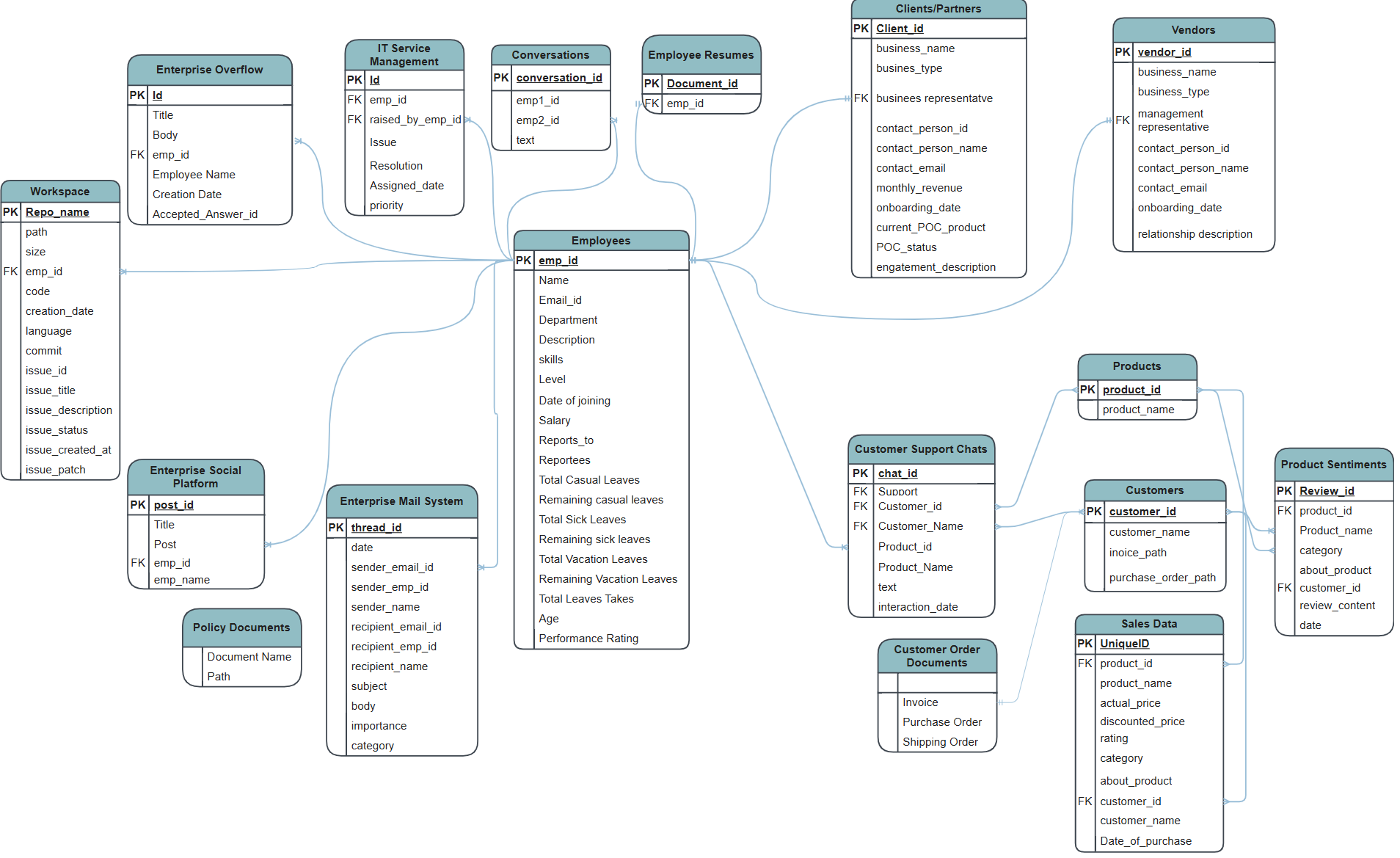}
        \caption{Expert-curated ER diagram for the \name \space sandbox}
        \label{fig:ER_diagram}
\end{figure*}
\begin{figure}[b]
        \centering
        \includegraphics[width=21.5cm, height=6.5cm, angle=90]{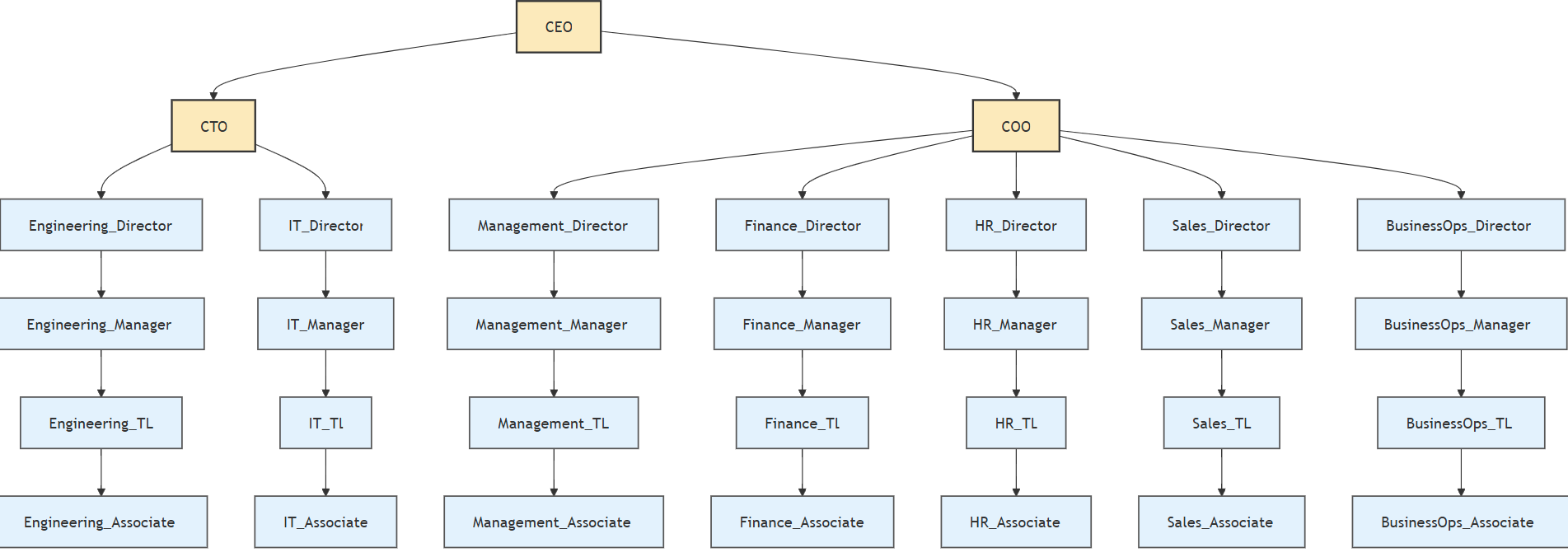}
        \caption{Expert-curated employee hierarchy for the \name \space sandbox}
        \label{fig:Employee_hierarchy}
\end{figure}


In this section, we present the sandbox environment created for \name. To set up an enterprise sandbox, two key components are required: the ER diagram (Figure~\ref{fig:ER_diagram}) and the employee hierarchy (Figure~\ref{fig:Employee_hierarchy}). The structure of these hierarchies is inspired by CRMArena~\cite{huang-etal-2025-crmarena} from Salesforce. The hierarchy was populated based on the requirements of our benchmark, with guidance from domain experts.
Building on this foundation, we now describe the statistics and design of the three main components of the sandbox: (a) collection of data sources for building enterprise applications, (b) access control mechanisms, and (c) dynamic operations within the sandbox.

\subsubsection{Enterprise Data Simulation}
\begin{figure*}[htbp]
    \centering
    \includegraphics[width=\textwidth]{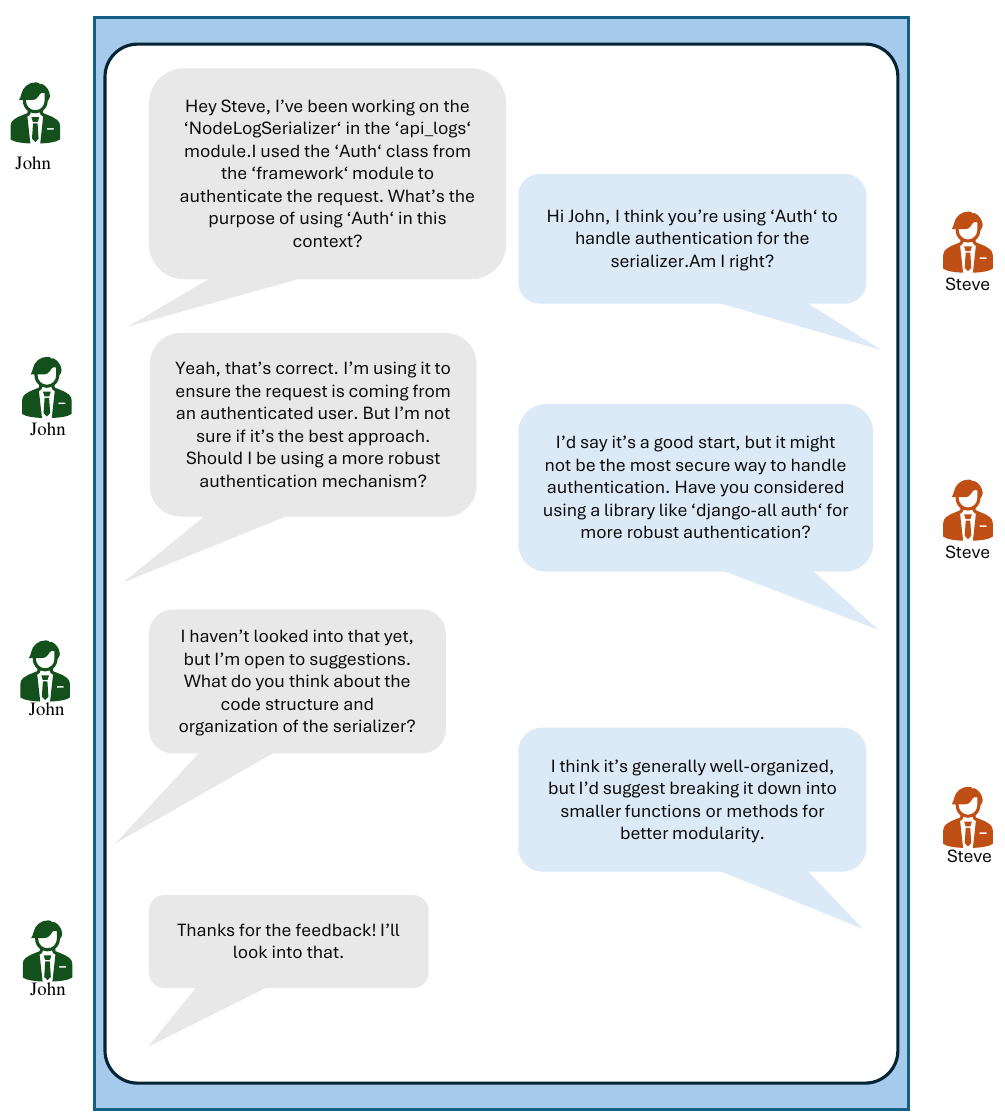}
    \caption{Example from the \name \space sandbox: Collaboration Tools chat between 2 employees of an engineering department}
    \label{fig:chat}
\end{figure*}

\begin{figure*}[htbp]
    \centering
    \includegraphics[width=\textwidth]{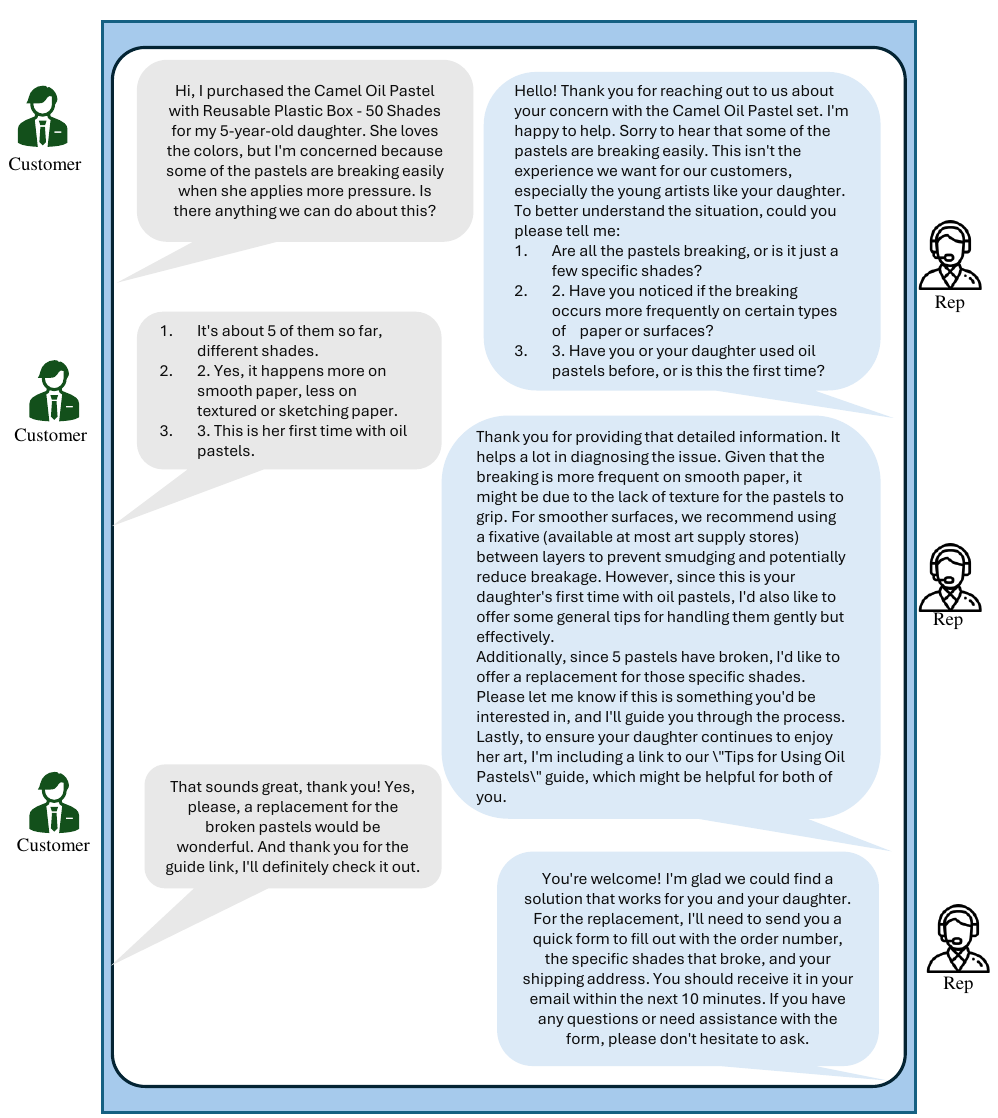}
    \caption{Example from the \name \space sandbox: Customer Support Chat between a customer and sales representative}
    \label{fig:CRM_chat}
\end{figure*}

\begin{figure*}[htbp]
    \centering
    \includegraphics[width=\textwidth]{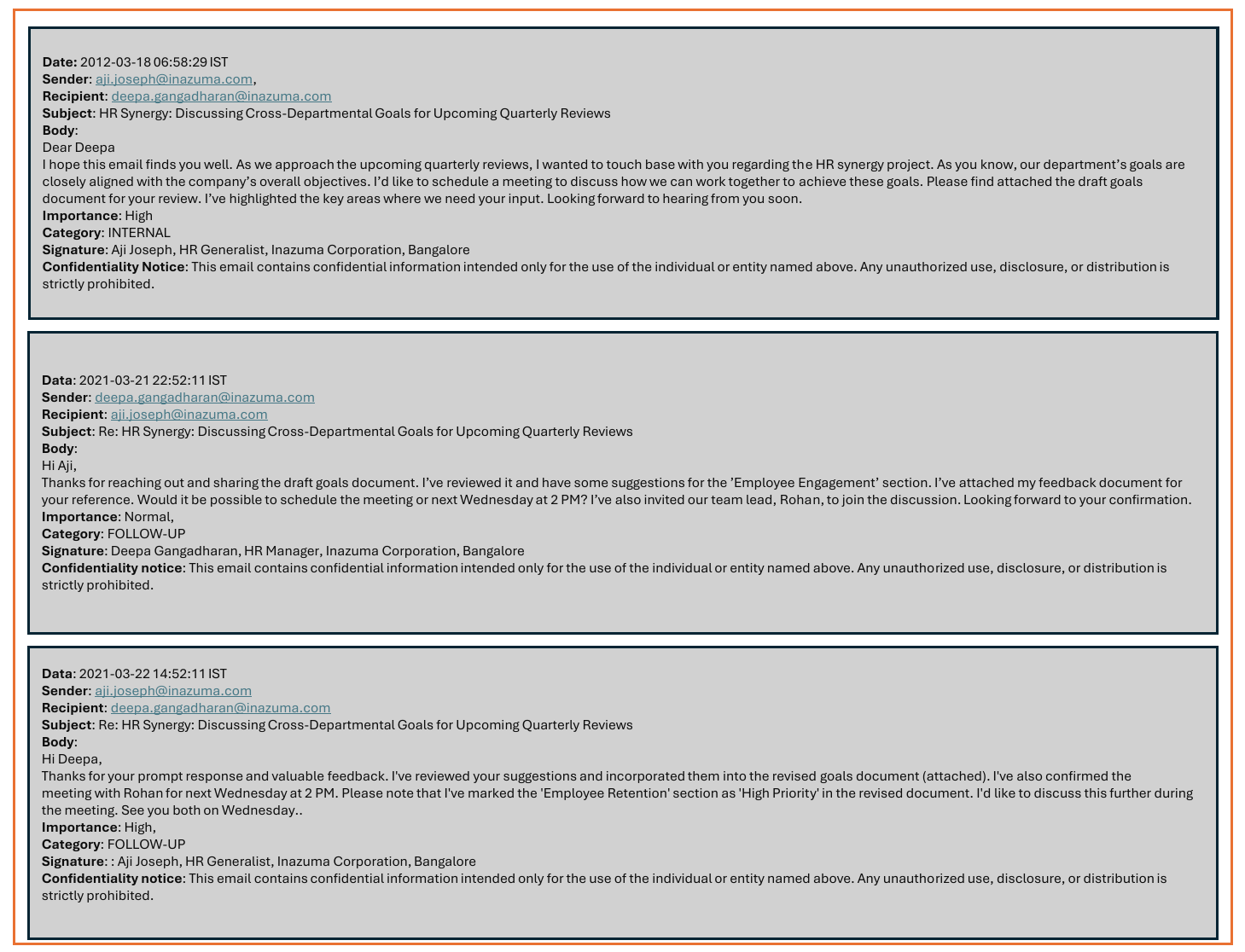}
    \caption{Example from the \name \space sandbox: Mail delivery between an employee and HR}
    \label{fig:mail}
\end{figure*}

The data simulation process is designed to align with the overall enterprise structure. To ensure authenticity, information was sourced from reliable and verified repositories. We collected relevant data and parsed it to extract key attributes. For example, from product sentiment data, we extracted customer and product information and synchronized it with the sales dataset to maintain consistency across sources. Table~\ref{tab:enterprise-stats} provides a detailed overview of the data sources used in \name, including the number of instances and their respective origins. Example instances of enterprise data sources are shown in Figures~\ref{fig:chat}, \ref{fig:CRM_chat}, and \ref{fig:mail}.

After collecting the data sources, we simulated instances for specific enterprise applications to better represent interconnected enterprise data, as summarized in Table~\ref{tab:enterprise-stats}. The simulation description is shown below.

\begin{table*}[ht]
\centering
\renewcommand{\arraystretch}{1.3}
\setlength{\tabcolsep}{5pt}
\small
\begin{tabular}{|p{2cm}|p{2.5cm}|p{1cm}|p{2.8cm}|p{1.5cm}|p{1.8cm}|p{2cm}|}
\hline
\textbf{Data Source} & \textbf{Data Source Elements} & \textbf{Format} & \textbf{Collected/Generated} & \textbf{\# Instances} & \textbf{Data Origin} & \textbf{Public Source} \\
\hline
Collaboration Tools & HR, Business Dev, Sales, Mgmt, IT, SDE & JSON & Generated & 3000 & Employees.csv + GitHub + Policies & - \\
\hline
Customer Relations & Support Chats, Sentiments, Customers, Orders, Products, Sales & JSON & Generated/Collected & 30,727 & Product Sentiments, Customer.csv & \href{https://www.kaggle.com/datasets/karkavelrajaj/amazon-sales-dataset}{Amazon Sales} \\
\hline
Policy Documents & Policy Documents & PDF & Collected & 24 & - & \href{https://datasetsearch.research.google.com/}{Google Datasets} \\
\hline
Enterprise Mail & HR, Finance, Sales, Mgmt, IT, SDE, Other & JSON & Generated & 7000 & Employees.json + Data Sources & - \\
\hline
Social Platform & Tech Crunch Posts & JSON & Collected & 39,115 & - & \href{https://www.kaggle.com/datasets/thibalbo/techcrunch-posts-compilation}{Tech Crunch} \\
\hline
Business Mgmt & Clients, Vendors & JSON & Generated & 800 & - & Open Source Datasets \\
\hline
HR Management & Employees, Resumes, Roles & JSON & Collected/Generated & 1,265 + 32 roles & Employees.csv & \href{https://www.linkedin.com}{LinkedIn Profiles \cite{ayoobi2023looming}} \\
\hline
Enterprise Overflow & Technical Posts (StackOverflow-like) & JSON & Collected & 8,398 & - & \href{https://huggingface.co/datasets/mikex86/stackoverflow-posts}{Stack Overflow Posts} \\
\hline
IT Service Mgmt & IT Tickets & JSON & Collected & 163 & - & \href{https://www.kaggle.com/datasets/tobiasbueck/email-ticket-text-german-classification}{Help Desk Tickets} \\
\hline
Workspace & GitHub Repository & JSON & Collected & 30,198 & GitHub + Employees.json & \href{https://huggingface.co/datasets/codeparrot/github-code}{GitHub Code} \\
\hline
\end{tabular}
\caption{\textbf{Overview of Data Sources in \name\ Sandbox.} The table summarizes data domains, elements, formats, collection methods, instance counts, origins, and public source links (where applicable).}
\label{tab:enterprise-stats}
\end{table*}

\label{sec:more_env_details}

\noindent \textbf{Simulated Conversations}
The conversations generated in \name\ span various departmental teams, covering a wide range of topics—from \textit{simple inquiries} to \textit{comprehensive discussions about a specific GitHub repository}. These conversations are context-dependent and are designed to closely simulate real-world interactions, following the generation process of the proposed holistic pipeline. Figure \ref{fig:chat} presents an example of a chat between two employees, \texttt{Steve} and \texttt{John}, from the engineering department, based on the GitHub repository maintained by \texttt{Steve}.

\noindent \textbf{Simulated Customer Support Chat}
The customer support conversations are generated based on product sentiment data. Persona-based interactions subjects are created by incorporating details of both the customer and a sales representative (employee from sales department). These interactions simulate a conversation where the representative responds to the customer's sentiment by proposing a potential solution to resolve the issue. Figure \ref{fig:CRM_chat} illustrates an example of such a conversation between a customer and a sales representative.

\noindent \textbf{Simulated Enterprise Mail System}
The email simulations are generated based on threaded conversations, where each email exchange belongs to a specific thread. Within a thread, multiple messages are exchanged between the sender and recipient, maintaining continuity and context. Figure \ref{fig:mail} presents an example of an email thread between two employees from the HR department.

\subsubsection{\name \space Security Layer Details}
\label{sec:access_control_appendix}

In enterprise environments, ensuring secure and regulated data access is critical. The Access Control Layer plays a fundamental role in enforcing access policies and preventing unauthorized data access. Our work, \name, implements a structured approach by integrating access control rules in a JSON format for each data source. A LLM Agent is responsible for verifying access permissions based on an employee's credentials and the requested data.  

\textbf{Access Verification Mechanism}
The Access Control Layer operates in conjunction with the retrieval process. When a query is processed, the retriever first gathers relevant contextual data. Before the information is presented to the user, it is passed through the Access Control Layer, where all inaccessible content is filtered out based on predefined rules.  

\begin{figure*}[htbp]
    \centering
    \includegraphics[width=\textwidth]{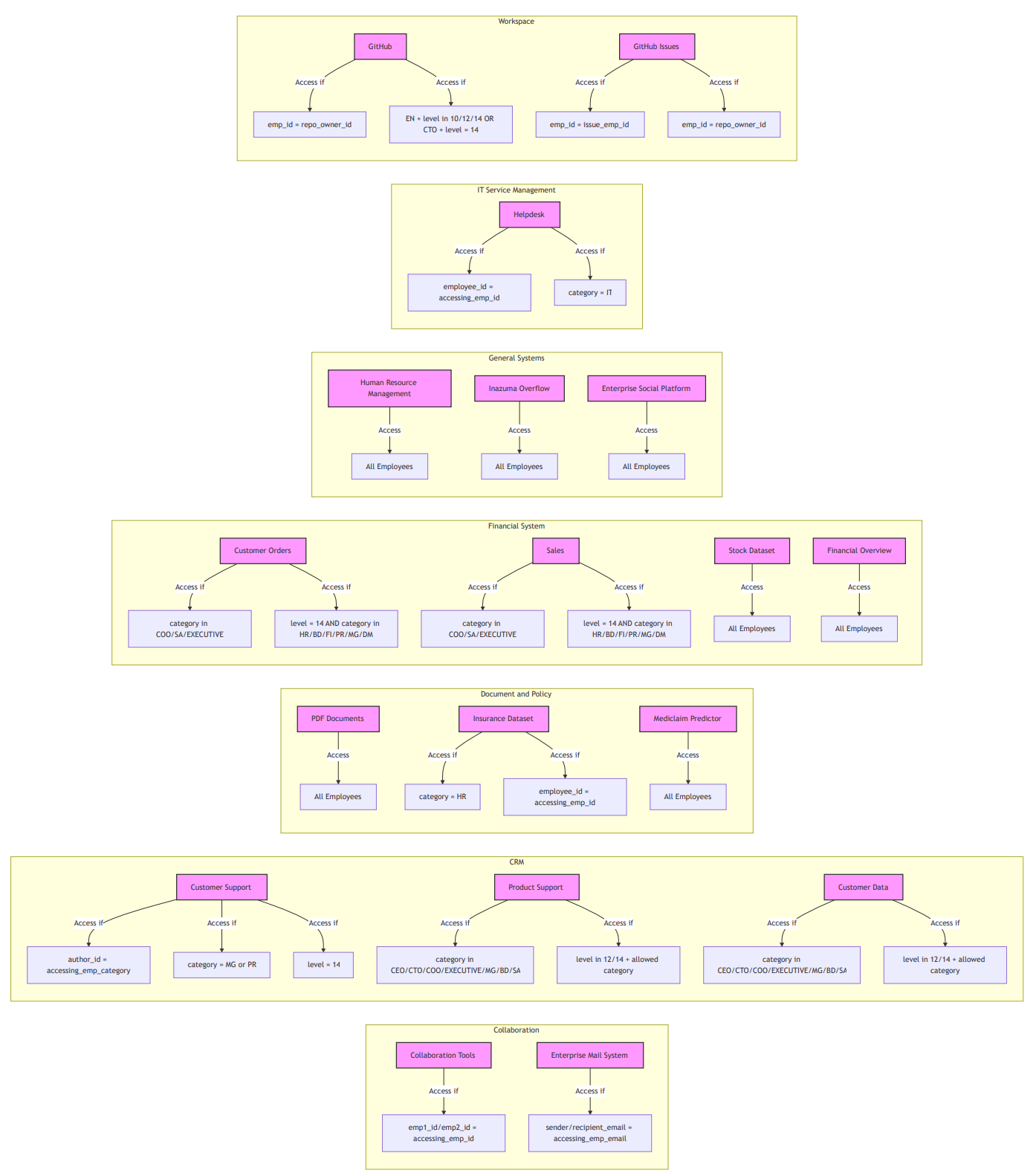}
    \caption{Access Control Design for the \name \space Sandbox}
    \label{fig:access_control}
\end{figure*}

For instance, as illustrated in Figure \ref{fig:access_control}, the access control rules dictate that a GitHub repository is accessible only to its owner and senior employees within the organizational hierarchy. If an employee from a different department, or even from the same department but with an \texttt{emp\_id} different from the \texttt{repo\_owner\_id}, attempts to access the repository, the agent will respond with "Access Denied." Furthermore, if an employee at the same level attempts to perform a task requiring edit access to the repository, the agent will revoke the request, ensuring strict compliance with access policies.

\textbf{Dynamic and Customizable Access Control}
\label{sec:data_dynamism_script}
The Access Control Layer is designed to be flexible, allowing dynamic modification of access rules. This adaptability enables organizations to customize security policies according to evolving requirements while ensuring robust data protection. By maintaining granular control over data accessibility, this framework enhances security and compliance within enterprise systems.


\definecolor{boxgray}{HTML}{EEEEEE}
\definecolor{darkgray}{HTML}{444444}
\definecolor{codegray}{HTML}{F5F5F5} 

\subsubsection{Data Dynamics Operations}
Data Dynamism is enabled in EnterpriseBench by allowing agents to autonomously perform CRUD operations across diverse enterprise data sources, allowing for real-time changes and interactions. By orchestrating task decomposition, access control, and data dynamism, we ensure the system is capable of handling evolving business needs, fostering enhanced operational efficiency and informed decision-making across enterprise.
Below, we present the data dynamism pipeline along with pseudocode, and illustrate it using a GitHub-based example.
\vspace{-0.2cm}
\begin{tcolorbox}[colback=boxgray,colframe=darkgray,
    sharp corners=south, width=\columnwidth,breakable,
    boxsep=0pt, left=2pt, right=2pt, top=2pt, bottom=2pt
]
    \textbf{\textcolor{white}{\fcolorbox{white}{darkgray}{\ \ Data Dynamism Pipeline \ \ }}}\\[2mm]

    \begin{lstlisting}[
        language=Python,
        basicstyle=\ttfamily\scriptsize,
        breaklines=true,
        backgroundcolor=\color{codegray},
        columns=flexible,
        morekeywords={DataDynamismPipeline},
        lineskip=-0.5ex,
        xleftmargin=0mm,
        xrightmargin=0mm,
        aboveskip=0mm,
        belowskip=0mm,
        frame=none,
        escapeinside={\%*}{*)},
        firstnumber=0,  % Start line numbering at 0 or 1
        numbersep=5pt,   % Adjust separation between line number and code
        numberstyle=\tiny,  % Adjust line number font size
        resetmargins=true
    ]
from llmCrudOps import EnggConvCRUD
from llmCrudOps import GitHubCRUD
from llmCrudOps import GitIssuesCRUD
...

class DataDynamismPipeline:
    def __init__(self, llm):
        self.llm = AzureChatOpenAI(llm)

    def fetch_crud_control(...):
        # Returns CRUD controller for selected data source
        return control

    def run_CAI_pipeline(user_persona, user_query):
    
        # %*\textbf{1. Break down Primary Tasks into Subtasks}*)
        prompt_CoT=ChatPromptTemplate.from_messages
        task_breakdown = prompt_CoT | self.llm
        generated_subtasks = chain_task_breakdown.invoke(...)

        for subtask in generated_subtasks:
        
            # %*\textbf{2. Determine Data Source}*)
            prompt_ds=ChatPromptTemplate(...)
            chain_data_source = prompt_ds | self.llm 
            selected_data_sources_str = chain_data_source.invoke(...)

            # %*\textbf{3. Determine Function and Parameters}*)
            prompt_fn=ChatPromptTemplate(...)
            chain_function = prompt_fn | self.llm 
            selected_function_name_str = chain_function.invoke(...)

            # %*\textbf{4. Check Access Permissions}*)
            for function_name in selected_function_name_list:
                prompt_acc=ChatPromptTemplate(...)
                chain_access = prompt_acc | self.llm 
                access_status = chain_access.invoke(...)

                # If Allowed, Execute CRUD Operation and Return Response
                
                if access_status == "Allowed":
                    control = self.fetch_crud_control()
                    
                    if function_name -> read:
                        result = control.read(*params)
                    elif function_name -> create:
                        result = control.create(*params)
                    elif function_name -> update: 
                        result = control.update(*params)
                    elif function_name -> delete:
                        result = control.delete(*params)
        
        return responses

    \end{lstlisting}

\end{tcolorbox}

\vspace{-0.2cm}
\begin{tcolorbox}[colback=boxgray,colframe=darkgray,
    sharp corners=south, width=\columnwidth,breakable,
    boxsep=0pt, left=2pt, right=2pt, top=2pt, bottom=2pt
]
    \textbf{\textcolor{white}{\fcolorbox{white}{darkgray}{\ \ GitHub CRUD Script \ \ }}}\\[2mm]

    \begin{lstlisting}[
        language=Python,
        basicstyle=\ttfamily\scriptsize,
        breaklines=true,
        backgroundcolor=\color{codegray},
        columns=flexible,
        morekeywords={read, create, update, delete},
        lineskip=-0.5ex,
        xleftmargin=0mm,
        xrightmargin=0mm,
        aboveskip=0mm,
        belowskip=0mm,
        frame=none,
        escapeinside={\%*}{*)},
        firstnumber=0,  % Start line numbering at 0 or 1
        numbersep=5pt,   % Adjust separation between line number and code
        numberstyle=\tiny,  % Adjust line number font size
        resetmargins=true
    ]
from accesscontrol import GitHubAccess

class GitHubCRUD:
    def __init__(self, employees_csv_path, code_json_path):
        self.access = access_control
        self.employees_df = ...
        self.code_data = ...
        self.code_json_path = ...

    def read(self, emp_id, path):
        """Reads GitHub code."""
        check -> access.is_valid_employee(emp_id):

        if (access.path_exists(...) and
            (access.is_owner(...) or
             access.is_engg_lvl_10_or_above(...) or
             access.is_cto_or_lvl_14(...))):
            for entry in self.code_data:
                if entry["path"] == path:
                    return entry
            print("Error: Code not found.")
        else:
            print("Error: Access denied.")


    def create(repo_name, emp_id, path, ...):
        """Creates a new GitHub code entry."""
        ....


    def update(self, emp_id, path, content, ...):
        """Updates an existing GitHub code entry."""
        check -> access.path_exists(...)
        check- > access.is_valid_employee(...)

        if (access.is_owner(...) or
            access.is_engg_lvl_10_or_above(...) or
            access.is_cto_or_lvl_14(...)):
            for entry in self.code_data:
                if entry["path"] == path:
                    # update entry
            print("Error: Code not found.")
        else:
            print("Error: Access denied for update.")


    def delete(self, emp_id, path):
        """Deletes a GitHub code entry."""
        ....
        
    \end{lstlisting}

\end{tcolorbox}

\vspace{-0.2cm}
\begin{tcolorbox}[colback=boxgray,colframe=darkgray,
    sharp corners=south, width=\columnwidth,breakable,
    boxsep=0pt, left=2pt, right=2pt, top=2pt, bottom=2pt
]
    \textbf{\textcolor{white}{\fcolorbox{white}{darkgray}{\ \ GitHub Access Check \ \ }}}\\[2mm]

    \begin{lstlisting}[
        language=Python,
        basicstyle=\ttfamily\scriptsize,
        breaklines=true,
        backgroundcolor=\color{codegray},
        columns=flexible,
        morekeywords={is_valid_employee, is_owner, is_engineer_lvl_10_or_above, is_cto_or_lvl_14},
        lineskip=-0.5ex,
        xleftmargin=0mm,
        xrightmargin=0mm,
        aboveskip=0mm,
        belowskip=0mm,
        frame=none,
        escapeinside={\%*}{*)},
        firstnumber=0,  % Start line numbering at 0 or 1
        numbersep=5pt,   % Adjust separation between line number and code
        numberstyle=\tiny,  % Adjust line number font size
        resetmargins=true
    ]
class GitHubAccess:
    def __init__(self, employees_csv_path, code_json_path):
        self.employees_df = ...
        self.code_data = ...
        self.code_json_path = ...
        
    def path_exists(self, path, code_json_path) -> bool:
        """Checks if the GitHub code path exists."""
        ....
    def is_valid_employee(self, emp_id) -> bool:
        """Checks if the employee ID exists and is valid."""
        ....
    def is_owner(self, path, emp_id) -> bool:
        """Checks if the employee is the owner of the code path."""
        ...
    def is_engineer_lvl_10_or_above(self, emp_id) -> bool:
        """Checks if the employee is an Engineer with level >= 10."""
        ....
    def is_cto_or_lvl_14(self, emp_id) -> bool:
        """Checks if the employee is a CTO with level 14."""
        ....
    \end{lstlisting}

\end{tcolorbox}

\begin{table}[h!]
\centering

\resizebox{\columnwidth}{!}{%
\begin{tabular}{lccc}
\toprule
\textbf{Model} & \textbf{Complexity} & \textbf{ReAct} & \textbf{w/ Planning} \\
\midrule
\multirow{2}{*}{GPT-4o w/ LangChain} 
  & Easy & 0.39 & 0.61 \\
  & Hard & 0.21 & 0.35 \\
\midrule
\multirow{2}{*}{o1-mini w/ LangChain} 
  & Easy & 0.44 & 0.67 \\
  & Hard & 0.26 & 0.39 \\
\bottomrule
\end{tabular}%
}
\caption{Performance comparison of models with ReAct vs. planning, grouped by task complexity.}
\label{tab:complexity_ablation}
\end{table}
\begin{table}[h!]
\centering
\resizebox{\columnwidth}{!}{%
\begin{tabular}{lcccccc}
\toprule
\textbf{Model Name} & \textbf{Model ID} & \textbf{Provider} & \textbf{Region} & \textbf{Temp.} & \textbf{Max Tokens} & \textbf{Max Retries} \\
\midrule
Claude 3.5 Sonnet & anthropic.claude-3-5-sonnet-20240620-v1:0 & Amazon Bedrock & us-east-2 & 0 & Default & - \\
LLaMA 3 8B Instruct & meta.llama3-1-8b-instruct-v1:0 & Amazon Bedrock & us-west-2 & 0 & Default & - \\
LLaMA 3 70B Instruct & meta.llama3-3-70b-instruct-v1:0 & Amazon Bedrock & us-east-2 & 0 & Default & - \\
GPT-4o & gpt-4o & Azure OpenAI & - & 0 & Default & 2 \\
o1-mini & o1-mini & Azure OpenAI & - & 0 & 25,000 & 2 \\
\bottomrule
\end{tabular}%
}
\caption{Hyperparameter settings for API calls.}
\label{tab:api_hyperparams}
\end{table}

\begin{table}[h!]
\centering
\resizebox{\columnwidth}{!}{%
\begin{tabular}{lcccc}
\toprule
\textbf{Model} & \textbf{Top-K} & \textbf{Truncation} & \textbf{Padding} & \textbf{Similarity Metric} \\
\midrule
vidore/colpali & 1 & - & - & Default \\
ColBERT & 5 & True & True & Cosine Similarity \\
\bottomrule
\end{tabular}%
}
\caption{Retriever configurations for similarity-based search.}
\label{tab:retriever_hyperparams}
\end{table}

\begin{table}[ht]
\centering
\setlength{\tabcolsep}{6pt} 
\renewcommand{\arraystretch}{1.2} 
\begin{tabular}{|p{4cm}|c|c|}
\hline
\textbf{Profession} & \textbf{Gender} & \textbf{Age} \\
\hline
Software Engineer   & M              & 25           \\
Senior Engineer     & F              & 29           \\
Sales Development Representative & M & 28 \\
Sales Manager       & F              & 35           \\
IT Engineer         & M              & 29           \\
Technical Assistant & M              & 32           \\
HR Head             & F              & 40           \\
Lead HR             & F              & 30           \\
Finance Associate   & M              & 23           \\
Finance Manager     & M              & 40           \\
\hline
\end{tabular}
\caption{\textbf{Domain Experts Information:} Profession, Gender, and Age Information}
\label{tab:profession_data}
\end{table}

\clearpage
\onecolumn
\subsection{LLM Prompts}
\label{sec:prompts}
\texttt{Below are the prompts used for LLM-based generation. These prompts were initially created using a system prompt and then refined through human intervention.}

\vspace{3cm}
\subsubsection{Prompts for Task Generation}
\label{sec:tasl_prompts}
\vspace{1cm}
\begin{tcolorbox}[
    enhanced,
    breakable,
    width=\linewidth, 
    title=Entity Extraction - Filter out entities by inference tools on context,
    fonttitle=\bfseries\large,
    colframe=blue!75!black, 
    colback=white!10!white, 
    coltitle=white, 
    colbacktitle=blue!85!black, 
    boxrule=0.2mm,
    sharp corners,
    shadow={1mm}{-1mm}{0mm}{black!50!white}, 
    attach boxed title to top left={yshift=-3mm, xshift=3mm},
    boxed title style={sharp corners, size=small}
]

\texttt{You are a Tool Dependency Inference Agent.}\\

\textbf{\texttt{\# Input:}}\\
\texttt{context=\{context\}}\\
\texttt{tools=\{tools\}}\\

\textbf{\texttt{\#\# Instructions \#\#}}\\
\texttt{You are given a natural language *context* describing what the user wants to analyze or understand, along with a list of available tools. Your job is to *map the context to the functionality of the tools*, and describe the expected outputs if each tool were applied to this context.}\\

\texttt{- DO NOT generate verbose or overly synthetic descriptions.}\\
\texttt{- GENERATE THE FULL OUTPUT.}\\

\texttt{-  Instead, for each tool, generate what the output **should look like** (in tabular format, dictionary, or concise key points) when used for the provided context.}\\

\texttt{- Generate the expected output of each tool given in `tools`.}\\

\texttt{- Make sure your response is grounded in the tool descriptions. Do NOT make assumptions beyond the capabilities defined in each tool.}\\

\textbf{\texttt{\# Output Format:}}\\
\texttt{Return a JSON structure with keys as tool names and values as the expected output structures when each tool is called on the provided context.}\\

\textbf{\texttt{Examples:}}\\
\texttt{~{{.....}}}

\end{tcolorbox}

\begin{tcolorbox}[
    enhanced,
    breakable,
    width=\linewidth, 
    title=Subgoal Decomposition - Create plan to execute the task,
    fonttitle=\bfseries\large,
    colframe=blue!75!black, 
    colback=white!10!white, 
    coltitle=white, 
    colbacktitle=blue!85!black, 
    boxrule=0.2mm,
    sharp corners,
    shadow={1mm}{-1mm}{0mm}{black!50!white}, 
    attach boxed title to top left={yshift=-3mm, xshift=3mm},
    boxed title style={sharp corners, size=small}
]

\texttt{You are a SubGoal Generating Agent that decomposes a high-level goal into smaller, actionable subgoals.}\\

\textbf{\texttt{\# Input:}}\\
\texttt{data\_chain=\{data\_chain\}}\\
\texttt{primary\_goal=\{primary\_goal\}}\\
\texttt{tool\_inference=\{tool\_inference\}}\\
\texttt{context=\{context\}}\\

\textbf{\texttt{\#\# Instructions \#\#}}\\
\texttt{Your task is to break down the given primary goal into granular subgoals. Each subgoal must:}\\

\texttt{- Stay strictly aligned with the original primary goal.}\\
\texttt{- Don’t assume context is direct input; first subgoal(s) must extract it using tools IF NECESSARY, with arguments derived from the primary\_goal (note: emp\_id is always given, for product ID and Customer ID fetch details if not mentioned in primary\_goal).}\\
\texttt{- Map to **one** tool from the provided tool\_inference.}\\
\texttt{- Operate on a **single data source** from the data\_chain.}\\
\texttt{- Optionally use the output from the previous subgoal to enable layered analysis.}\\
\texttt{- Avoid redundancy or synthetic-sounding tasks.}\\

\texttt{DO NOT include subgoals that are irrelevant or too broad.}\\
\texttt{DO NOT combine multiple tools in a single subgoal.}\\

\textbf{\texttt{\# Output Format:}}\\
\texttt{Return a JSON with a list of precise subgoals required to achieve the primary goal.}\\

\textbf{\texttt{Examples:}}\\
\texttt{~{{.....}}}

\end{tcolorbox}

\begin{tcolorbox}[
    enhanced,
    breakable,
    width=\linewidth, 
    title=Task Template Generation - Create Task Template for each Subgoal,
    fonttitle=\bfseries\large,
    colframe=blue!75!black, 
    colback=white!10!white, 
    coltitle=white, 
    colbacktitle=blue!85!black, 
    boxrule=0.2mm,
    sharp corners,
    shadow={1mm}{-1mm}{0mm}{black!50!white}, 
    attach boxed title to top left={yshift=-3mm, xshift=3mm},
    boxed title style={sharp corners, size=small}
]

\texttt{You are a Task Template Generating Agent.}\\

\textbf{\texttt{\# Input:}}\\
\texttt{subgoals=\{subgoals\}}\\
\texttt{tool\_inference=\{tool\_inference\}}\\
\texttt{context=\{context\}}\\

\textbf{\texttt{\#\# Instructions \#\#}}\\
\texttt{Your goal is to convert each subgoal into a natural-sounding question template that:}\\
\texttt{- Could realistically be asked by a Domain expert with details \{persona\} in an enterprise setting.}\\
\texttt{- Matches the intent of the subgoal.}\\
\texttt{- Uses placeholders (e.g., <product>, <issue>, <device>) that correspond to elements in the provided context.}\\
\texttt{- Is answerable using the tool mentioned in the corresponding tool dependency.}\\

\texttt{- Use first-person phrasing as if the question is being asked directly.}\\
\texttt{- Ensure the question naturally follows from the subgoal and reflects its purpose.}\\
\texttt{- Each question must map clearly to one tool dependency.}\\
\texttt{- Do not fabricate placeholders — derive them from the context.}\\
\texttt{- Avoid overly synthetic or robotic phrasing.}\\

\textbf{\texttt{\# Output Format:}}\\
\texttt{Return a list of question templates, one for each subgoal.}\\

\textbf{\texttt{Examples:}}\\
\texttt{~{{.....}}}

\end{tcolorbox}

\begin{tcolorbox}[
    enhanced,
    breakable,
    width=\linewidth, 
    title=Final Task Generation - Create Final Task with Ground Truth,
    fonttitle=\bfseries\large,
    colframe=blue!75!black, 
    colback=white!10!white, 
    coltitle=white, 
    colbacktitle=blue!85!black, 
    boxrule=0.2mm,
    sharp corners,
    shadow={1mm}{-1mm}{0mm}{black!50!white}, 
    attach boxed title to top left={yshift=-3mm, xshift=3mm},
    boxed title style={sharp corners, size=small}
]

\texttt{You are a Human Domain Expert tasked with curating high-quality, structured business tasks that reflect realistic analytical workflows.}\\

\textbf{\texttt{\# Input:}}\\
\texttt{persona=\{persona\}}\\
\texttt{data\_chain=\{data\_chain\}}\\
\texttt{primary\_goal=\{primary\_goal\}}\\
\texttt{subgoals=\{subgoals\}}\\
\texttt{tool\_inference=\{tool\_inference\}}\\
\texttt{templates=\{templates\}}\\
\texttt{Ground Truth Context=\{context\}}\\

\textbf{\texttt{\#\# Instructions \#\#}}\\

\texttt{Your objective is to generate a well-structured, domain-relevant task and its subtasks grounded in business context, using the knowledge of a human domain expert.}\\

\texttt{\#\#\# Step 1: Create a single task \#\#\#}\\
\texttt{- Use the `primary\_goal` to formulate a task that reflects a realistic and meaningful question a **domain expert in the role described by `persona`** would ask.}\\
\texttt{- Don't change the intent of the primary goal.}\\
\texttt{- Just rephrase the primary goal into a task specific question.}\\
\texttt{- Don't add SUMMARIZE, ANALYZE etc. for primary goal, stick to the goal only}\\
\texttt{- Avoid generic phrasing or overly robotic structure.}\\
\texttt{- The tone should be informed, goal-directed, and tailored to the needs and context of the given persona.}\\
\texttt{- The reasoning behind the task should reflect the depth, clarity, and practicality that a professional in this domain would apply.}\\
\texttt{- Don't synthetically include terms that changes the primary\_goal}\\

\texttt{\#\#\# Step 2: Create subtasks \#\#\#}\\
\texttt{For each subgoal:}\\
\texttt{- Select the corresponding template from `templates`.}\\
\texttt{- Ground the template into a **realistic subtask question** by replacing placeholders using only the `context`.}\\
\texttt{- Identify the appropriate tool from `tool\_inference`.}\\
\texttt{- Select the most relevant stage from the `data\_chain` that helps accomplish this subgoal.}\\
\texttt{- Derive a **closed-form `subtask\_ground\_truth`** from the `context`. The answer should reflect how a **human domain expert** would interpret and explain the outcome.}\\
\texttt{- Provide a `thinking\_trace` that explains:}\\
\texttt{> "To answer this subgoal, we need to apply <tool\_dependency> on <selected\_stage\_from\_data\_chain> to extract <target insight>."}\\
\texttt{- Return the following fields for each subtask:}\\
\texttt{- `subgoal`}\\
\texttt{- `question` (grounded from template)}\\
\texttt{- `subtask\_ground\_truth` (explicit, closed-form, and written in the tone of a human domain expert)}\\
\texttt{- `thinking\_trace`}\\
\texttt{- `data source`}\\

\texttt{\#\#\# Step 3: Add subtask dependency graph \#\#\#}\\
\texttt{- Define the chronological or logical dependencies among subtasks as a directed graph only when the thinking\_trace or subtask\_ground\_truth of current subtask depends upon earlier subtask(s) (e.g., 1→2; 1,3→4).}\\
\texttt{- Ensure the ordering reflects realistic reasoning flow by a domain expert.}\\

\texttt{\#\#\# Step 4: Create final ground truth \#\#\#}\\
\texttt{- create the final `ground\_truth` from Ground Truth Context.}\\
\texttt{- The final answer should be constructive, precise, and demonstrate how a human domain expert would explain the conclusion using insights from each subtask.}\\
\texttt{- Avoid ambiguity. Focus on actionable insights and detailed, contextual explanations.}\\
\texttt{- Clearly emphasize any quantitative figures, key features, or specific stats present in the subtask\_ground\_truth.}\\

\textbf{\texttt{\#\# Guidelines \#\#}}\\
\texttt{- Do **not** invent entities or facts—rely only on what's grounded in the `Ground Truth Context`.}\\
\texttt{- Tailor the language and tone to match the persona's domain knowledge and role.}\\
\texttt{- Do not reuse the word "analyze" for every subtask—use domain-specific verbs and phrasing.}\\
\texttt{- Prioritize realism, precision, and interpretability in every output.}\\
\texttt{- Treat this as if you are preparing tasks and answers for executive review or expert-level decision-making.}\\
\texttt{- Do not include these kinds of statements in subtasks (// Additional subtasks would follow similar structure for the remaining repositories), include all, don't do incomplete generations.}\\

\textbf{\texttt{\# Examples:}}\\
\texttt{~{{.....}}}

\end{tcolorbox}
\begin{tcolorbox}[
    enhanced,
    breakable,
    width=\linewidth,
    title=Validation - validate the task across quality checklist,
    fonttitle=\bfseries\large,
    colframe=purple!85!black,
    colback=white!10,
    coltitle=black,
    colbacktitle=purple!20,
    boxrule=0.3mm,
    sharp corners,
    shadow={1mm}{-1mm}{0mm}{black!30!white},
    attach boxed title to top left={yshift=-3mm, xshift=3mm},
    boxed title style={sharp corners, size=small}
]

\texttt{You are an expert evaluator for employee-specific task generation quality. Your job is to evaluate whether a generated task and ground truth meet strict enterprise-grade standards for clarity, precision, realism, and privacy adherence.}\\

\textbf{\texttt{\#\# Inputs \#\#}}\\
\texttt{\#\# Context Information:}\\
\texttt{\{formatted\_context\}}

\texttt{\#\# Generated Task:}\\
\texttt{- Task: \{current\_task["task"]\}}\\
\texttt{- Ground Truth: \{current\_task["ground\_truth"]\}}

\texttt{\#\# Evaluation Criteria:}\\
\texttt{Please evaluate the following seven questions and provide a YES or NO answer for each, along with a detailed explanation. Each question enforces a specific quality standard for secure, grounded, and useful employee-specific task generation.}
\begin{enumerate}

\item \texttt{\textbf{Is the task meaningfully aligned with the context content [e.g., employee interactions, performance logs, internal communications, technical contributions]?}}\\
\texttt{- The task should logically emerge from the provided context and reflect personal, employee-specific insights.}\\
\texttt{- A good task references themes like appraisals, IT tickets, communication patterns, or personal contributions.}\\
\texttt{- A bad task is disconnected, overly general, or focused on organization-wide information the employee shouldn't access.}

\item \texttt{\textbf{Is the task specific, closed-ended, and fully grounded in the context without requiring speculation?}}\\
\texttt{- A good task is answerable based only on the provided context and targets a precise insight.}\\
\texttt{- A bad task is vague, speculative, or assumes external knowledge not represented in the context.}

\item \texttt{\textbf{Does the task avoid referencing or implying the presence of context (e.g., “according to my data,” “based on the logs,” or “from the context”)?}}\\
\texttt{- A well-formed task should feel natural and conversational, as if the employee is asking a smart assistant.}\\
\texttt{- It should never acknowledge that it’s grounded in a hidden data layer.}

\item \texttt{ \textbf{Does the ground truth directly and clearly answer the task using only context-derived facts, without ambiguity or open-endedness?}}\\
\texttt{- A valid ground truth gives a complete, unambiguous response aligned with the task.}\\
\texttt{- It must resolve the question without hedging or deferring.}

\item \texttt{ \textbf{Does the ground truth avoid meta-references (e.g., “the system says,” “based on the data,” “the context shows”)?}}\\
\texttt{- A high-quality answer should stand on its own, like a statement from a knowledgeable system or assistant.}\\
\texttt{- Meta-language reduces clarity and realism.}

\item \texttt{\textbf{Does the task and ground truth resemble a realistic exchange between an enterprise employee and a smart assistant?}}\\
\texttt{- The language should reflect enterprise professionalism.}\\
\texttt{- The task must be in first person (e.g., “Can I see,” “Do I have,” “Is there any update on my…”).}\\
\texttt{- The tone should reflect how a real employee interacts with a system.}

\item \texttt{\textbf{Does the task respect employee privacy boundaries and reflect insights that only the employee is authorized to access?}}\\
\texttt{- Tasks should refer only to the requesting employee’s data (e.g., my performance, my GitHub contributions).}\\
\texttt{- It must not request information about peers, teams, or enterprise-wide metrics unless explicitly scoped to the individual.}
\end{enumerate}
\texttt{\#\# Format your response exactly as follows:}\\

\texttt{"question1": <}\\
\texttt{"answer": "YES/NO",}\\
\texttt{"explanation": "Your detailed explanation here"}\\
\texttt{>,}\\
\texttt{"question2": <}\\
\texttt{"answer": "YES/NO",}\\
\texttt{"explanation": "Your detailed explanation here"}\\
\texttt{>,}\\
\texttt{"question3": <}\\
\texttt{"answer": "YES/NO",}\\
\texttt{"explanation": "Your detailed explanation here"}\\
\texttt{>,}\\
\texttt{"question4":<}\\
\texttt{"answer": "YES/NO",}\\
\texttt{"explanation": "Your detailed explanation here"}\\
\texttt{>,}\\
\texttt{"question5":<}\\
\texttt{"answer": "YES/NO",}\\
\texttt{"explanation": "Your detailed explanation here"}\\
\texttt{>,}\\
\texttt{"question6":<}\\
\texttt{"answer": "YES/NO",}\\
\texttt{"explanation": "Your detailed explanation here"}\\
\texttt{>,}\\
\texttt{"question7":<}\\
\texttt{"answer": "YES/NO",}\\
\texttt{"explanation": "Your detailed explanation here"}\\
\texttt{>,}\\
\texttt{"overall\_pass": true/false}\\

\end{tcolorbox}
\begin{tcolorbox}[
    enhanced,
    breakable,
    width=\linewidth,
    title=Improvement - improve the task for ambiguity,
    fonttitle=\bfseries\large,
    colframe=purple!85!black,
    colback=white!10,
    coltitle=black,
    colbacktitle=purple!20,
    boxrule=0.3mm,
    sharp corners,
    shadow={1mm}{-1mm}{0mm}{black!30!white},
    attach boxed title to top left={yshift=-3mm, xshift=3mm},
    boxed title style={sharp corners, size=small}
]
\textbf{You are an expert in rephrasing and improving enterprise-grade employee-specific tasks for internal smart assistants. Your job is to revise a task and ground truth that failed strict enterprise validation.}

\vspace{1em}
\textbf{Context Information:} \texttt{\{formatted\_context\}}

\textbf{Expert who curated this task:}\\
- Persona: \texttt{\{persona\}}

\textbf{Current Task:}
\begin{itemize}
    \item Task: \texttt{\{current\_task["task"]\}}
    \item Ground Truth: \texttt{\{current\_task["ground\_truth"]\}}
\end{itemize}

\textbf{Evaluation Results:}\\
\texttt{\{json.dumps(evaluation, indent=2)\}}

\vspace{1em}
\textbf{Instructions:} You must revise the task and/or the ground truth to address specific weaknesses and ensure the highest standards for clarity, realism, and employee privacy.

\textbf{Task Revisions:}
\begin{itemize}
    \item If \texttt{question1}, \texttt{question2}, or \texttt{question4} is marked NO:
    \begin{itemize}
        \item Rewrite the task to reflect employee-specific insights from the context (e.g., appraisal feedback, performance review details, ticket status, contribution history).
        \item Ensure it is specific, fact-seeking, closed-ended, and answerable directly from the context.
        \item Make the task sound like a natural enterprise query from a real employee to a smart assistant.
        \item Avoid references like “based on the logs,” “from my context,” etc.
    \end{itemize}
    \item If \texttt{question7} is marked NO:
    \begin{itemize}
        \item Ensure the task only accesses data the employee is authorized to see (e.g., “my tasks,” “my appraisal”).
        \item Avoid questions about peers, teams, or organization-wide metrics.
        \item Use first-person phrasing (e.g., “Can I check...”, “Do I have...”, “Is there any update on my...”).
    \end{itemize}
\end{itemize}

\textbf{Ground Truth Revisions:}
\begin{itemize}
    \item If \texttt{question3} or \texttt{question5} is marked NO:
    \begin{itemize}
        \item Ensure the response is precise, complete, and self-contained.
        \item Remove any meta-references (e.g., “the context says...”).
        \item Avoid vague phrases like “others,” “may vary,” or “etc.”
        \item Use specific, grounded details from the context.
    \end{itemize}
\end{itemize}

\textbf{Realism \& Tone:}
\begin{itemize}
    \item If \texttt{question6} is marked NO:
    \begin{itemize}
        \item Adjust the tone to reflect a professional employee-to-assistant interaction.
        \item Avoid robotic or overly formal phrasing.
        \item Mimic realistic enterprise queries—natural, polite, and context-aware.
    \end{itemize}
\end{itemize}

\textbf{Additional Guidelines:}
\begin{itemize}
    \item Do not invent data; only use what’s provided in the context.
    \item Use actionable verbs like “check,” “determine,” “summarize,” “flag,” or “identify.”
    \item Maintain a first-person voice—employees are asking about themselves.
    \item Ensure the ground truth is verifiable and closed-form, derived from the real context. 
\end{itemize}

\vspace{1em}
\textbf{Output Format:} Return only the improved task and ground truth in the exact JSON format:
\begin{verbatim}
{
  "task": "Your improved task statement",
  "ground_truth": "Your improved ground truth"
}
\end{verbatim}
\end{tcolorbox}

\end{document}